\documentclass[runningheads]{llncs}

\usepackage{eccv}
\usepackage[square, comma, sort&compress, numbers]{natbib}

\usepackage{eccvabbrv}

\usepackage{graphicx}
\usepackage{booktabs}

\usepackage[pagebackref,breaklinks,colorlinks,citecolor=eccvblue]{hyperref}

\usepackage{orcidlink}

\usepackage{epsfig}
\usepackage{graphicx}
\usepackage{amsmath}
\usepackage{amssymb}

\usepackage{graphicx}
\usepackage{booktabs}
\usepackage{bm}
\usepackage{caption}
\usepackage{multirow}

\usepackage{wrapfig}
\usepackage{epsfig}
\usepackage{graphicx}
\usepackage{makecell}
\usepackage{caption}
\usepackage{wrapfig}
\usepackage{booktabs}

\usepackage{pifont}

\usepackage{cuted}
\usepackage{capt-of}
\usepackage[capitalize]{cleveref}

\usepackage{overpic}
\usepackage{enumitem} 
\usepackage{overpic} 
\usepackage{color}

\definecolor{turquoise}{cmyk}{0.65,0,0.1,0.3}
\definecolor{purple}{rgb}{0.65,0,0.65}
\definecolor{dark_green}{rgb}{0, 0.5, 0}
\definecolor{orange}{rgb}{0.8, 0.6, 0.2}
\definecolor{red}{rgb}{0.8, 0.2, 0.2}
\definecolor{darkred}{rgb}{0.6, 0.1, 0.05}
\definecolor{blueish}{rgb}{0.0, 0.3, .6}
\definecolor{light_gray}{rgb}{0.7, 0.7, .7}
\definecolor{pink}{rgb}{1, 0, 1}
\definecolor{greyblue}{rgb}{0.25, 0.25, 1}

\newcommand\cnum[1]{\raisebox{.5pt}{\textcircled{\raisebox{-0.9pt}{#1}}}}

\usepackage{blindtext}

\renewcommand{\paragraph}[1]{\vspace{1em}\noindent\textbf{#1}.}

\begin{document}

\title{StructLDM: Structured Latent Diffusion for 3D Human Generation}    


\vspace{-0.2in}
\author{Tao Hu \and Fangzhou Hong \and Ziwei Liu}

\vspace{-0.28in}
\institute{S-Lab, Nanyang Technological University, Singapore}
\vspace{0.16in}

\maketitle

\newcommand{\nickname}{StructLDM}

\begin{figure}
    \centering
    \vspace{-20pt}
    \includegraphics[width=\textwidth]{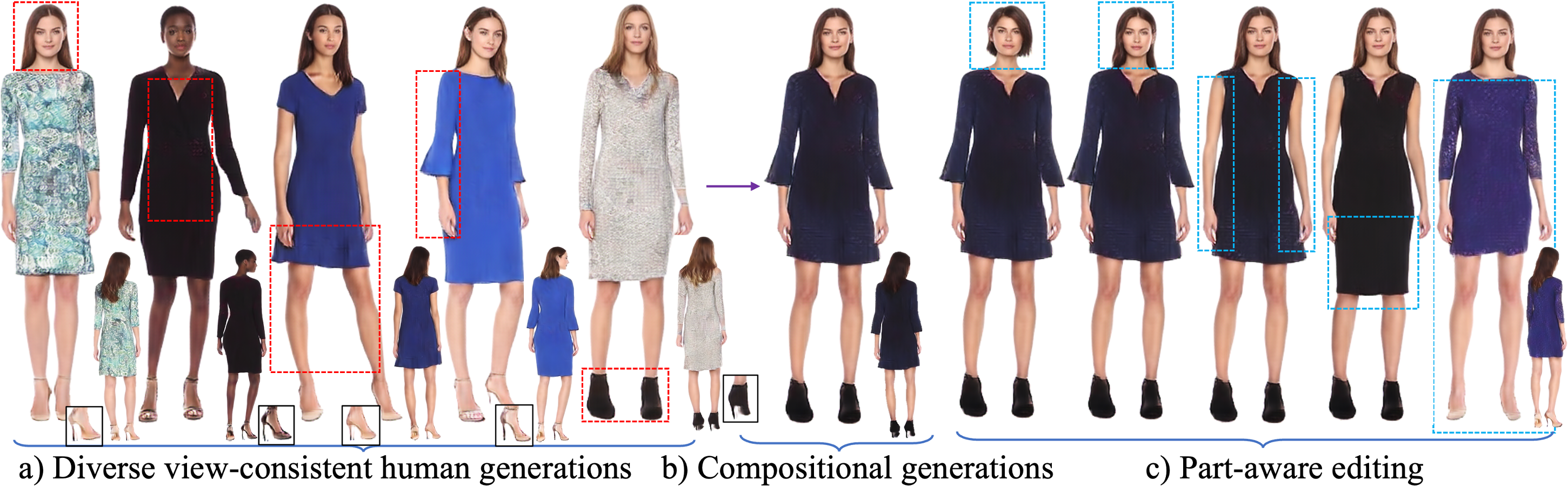}
    \vspace{-15pt}
    \captionof{figure}{\nickname{} generates diverse view-consistent humans, and supports different levels of controllable generations and editings, such as compositional generations by blending the five selected parts from a), and part-aware editings such as identity swapping, local clothing editing, 3D virtual try-on, etc. Note that the generations and editing are \textbf{clothing-agnostic} without clothing types or masks.}
    \vspace{-30pt}
    \label{fig:teaser}   
\end{figure}

\begin{abstract}
Recent 3D human generative models have achieved remarkable progress by learning 3D-aware GANs from 2D images. However, existing 3D human generative methods model humans in a compact 1D latent space, ignoring the articulated structure and semantics of human body topology. In this paper, we explore more expressive and higher-dimensional latent space for 3D human modeling and propose \textbf{\nickname}, a diffusion-based unconditional 3D human generative model, which is learned from 2D images. \textbf{\nickname} solves the challenges imposed due to the high-dimensional growth of latent space with three key designs: \textbf{1) A semantic structured latent space} defined on the dense surface manifold of a statistical human body template. \textbf{2) A structured 3D-aware auto-decoder} that factorizes the global latent space into several semantic body parts parameterized by a set of conditional structured local NeRFs anchored to the body template, which embeds the properties learned from the 2D training data and can be decoded to render view-consistent humans under different poses and clothing styles. \textbf{3) A structured latent diffusion model} for generative human appearance sampling. Extensive experiments validate \nickname's state-of-the-art generation performance and illustrate the expressiveness of the structured latent space over the well-adopted 1D latent space. Notably, \nickname{} enables different levels of controllable 3D human generation and editing, including pose/view/shape control, and high-level tasks including compositional generations, part-aware clothing editing, 3D virtual try-on, etc. Project page: {\footnotesize \href{https://taohuumd.github.io/projects/StructLDM/}{taohuumd.github.io/projects/StructLDM}}.


\end{abstract}
\section{Introduction}
\label{sec:intro}

Generating and editing high-quality 3D digital humans have been long-studied topics. It empowers many downstream applications, \eg, virtual try-on, and telepresence. Existing works use 3D-aware GAN to learn 3D human generation from 2D images~\cite{enarfgan,eva,avatargen,ag3d}, which suffer from low-fidelity generation. In this paper, we contribute a new paradigm of 3D human generation by proposing \textbf{\nickname{}}, a diffusion-based 3D human generation model with structured human representation that learns from 2D multi-view images or monocular videos. 

The problems of 3D-aware human GANs are arguably two parts. Firstly, existing works overlook the semantics and structure of the human body. They sample humans in a compact 1D space, which severely limits their controlling ability. Instead, we propose to explore higher-dimensional semantic latent space for human representation, which allows better capturing of fine details of 3D humans and easy local editing. Secondly, although the 3D-aware GAN has been a success in generating 3D faces and single-class instances~\cite{eg3d,stylesdf}, its application in 3D human generation has not achieved comparable generation quality. It shows the complexity of 3D human modeling over other subjects and indicates the need for a more powerful model to advance the field. Following the recent success in diffusion models~\cite{ddpm,ldm}, we bring the power of diffusion model to 3D human generation. 

Combining the powerful diffusion model and the structured latent representation, we achieve diverse and high-quality 3D human generation, as shown in Fig.~\ref{fig:teaser} a). We have also demonstrated novel editing results without using clothing masks, such as 3D compositional generation, clothing editing, and 3D virtual in Fig.~\ref{fig:teaser} b) and c). 

However, the extension over 1D latent space is non-trivial, which imposes more challenges on the generative models due to the high-dimensional growth of the global latent space. To tackle these challenges, in contrast to the well-adopted global 1D latent learning, we propose to model the human latent space locally by proposing three key designs: \textbf{1)} a structured and dense human representation; \textbf{2)} an auto-decoder architecture to embed latent features; \textbf{3)} a latent diffusion model with structure-specific normalization.

Firstly, to fully explore the rich semantics and articulations of the human body, we propose a structured and dense human representation semantically corresponding to human body mesh~\cite{smpl}. It preserves the articulated nature of the human body, and enables detailed appearance capture and editing, as shown in Fig. \ref{fig:framework}. In contrast to \cite{eg3d,stylepeople,Chen2022UVVF,Sun2022Next3DGN,enarf,eg3dhuman} that rely on an implicit mapping network (\eg,  StyleGAN \cite{karras2019style}) to map 1D embedding to latent space while sample humans in 1D space, we explicitly model the structured latent space with explainable differentiable rendering \textbf{without relying on mapping networks}, which faithfully preserves the fidelity and semantic structures of the latent or embedding space. 

Secondly, we design a structured 3D-aware auto-decoder that embeds the structured latent from the 2D training dataset to a shared latent space. We propose to divide the human body into several parts for rendering. At the core of the auto-decoder is a set of structured NeRFs that are locally conditioned on the structured latent space to render a specific body part. Both reconstruction and adversarial supervision are employed to encourage high-fidelity and high-quality image synthesis with high robustness to estimation errors of the training dataset (\eg, human pose or camera estimation errors).

Thirdly, with the structured latents prepared, we learn a latent diffusion model to sample in that space. Since the latents are structured and semantically aligned, we further tailor the diffusion process by using a structure-aligned normalization, which helps to better capture the distribution of our data. Together with the structured latent, the diffusion model enables different levels of controllable 3D human generation and editing, as shown in Fig. \ref{fig:teaser}. 

Quantitative and qualitative experiments are performed on three datasets with different setup: monocular videos of UBCFashion \cite{ubcfashion}, multi-view images of RenderPeople \cite{renderpeople} and THUman2.0~\cite{thuman2}. They illustrate the versatility and scalability of \nickname{}. To conclude, our contributions are listed as follows.

\noindent\textbf{1)} Different from the widely adopted 1D latent, we explore the higher-dimensional latent space \textbf{without latent mapping} for 3D human generation and editing.

\noindent\textbf{2)} We propose \nickname{}, a diffusion-based 3D human generative model, which achieves state-of-the-art results in 3D human generation.

\noindent\textbf{3)} Emerging from our design choices, we show novel controllable generation and editing tasks, \eg, 3D compositional generations, part-aware 3D editing, 3D virtual try-on.

\newcommand{\tabAblation}{
\begin{table*}[t]
    \vspace{-0.1in}
    \begin{minipage}{.45\linewidth}
        \begin{center}  
            \includegraphics[width=\linewidth]{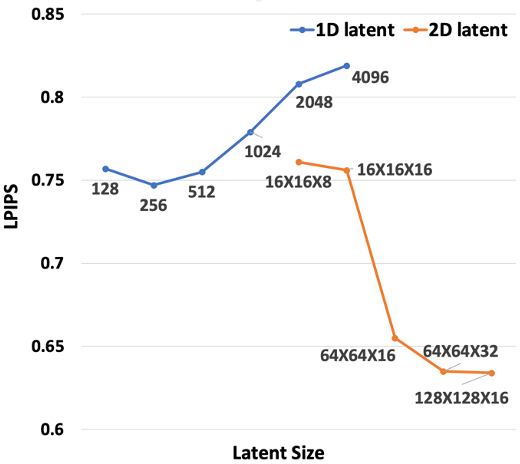}
        \end{center}            
        \vspace{-0.1in}
        \captionof{figure}{Ablation of latent size in auto-decoder.}
        \vspace{-0.06in}
        \label{fig:ab_ad_quan_latent}
    \end{minipage} \hfill  
	\begin{minipage}{.5\linewidth}
		\begin{center} 
			\includegraphics[width=\linewidth]{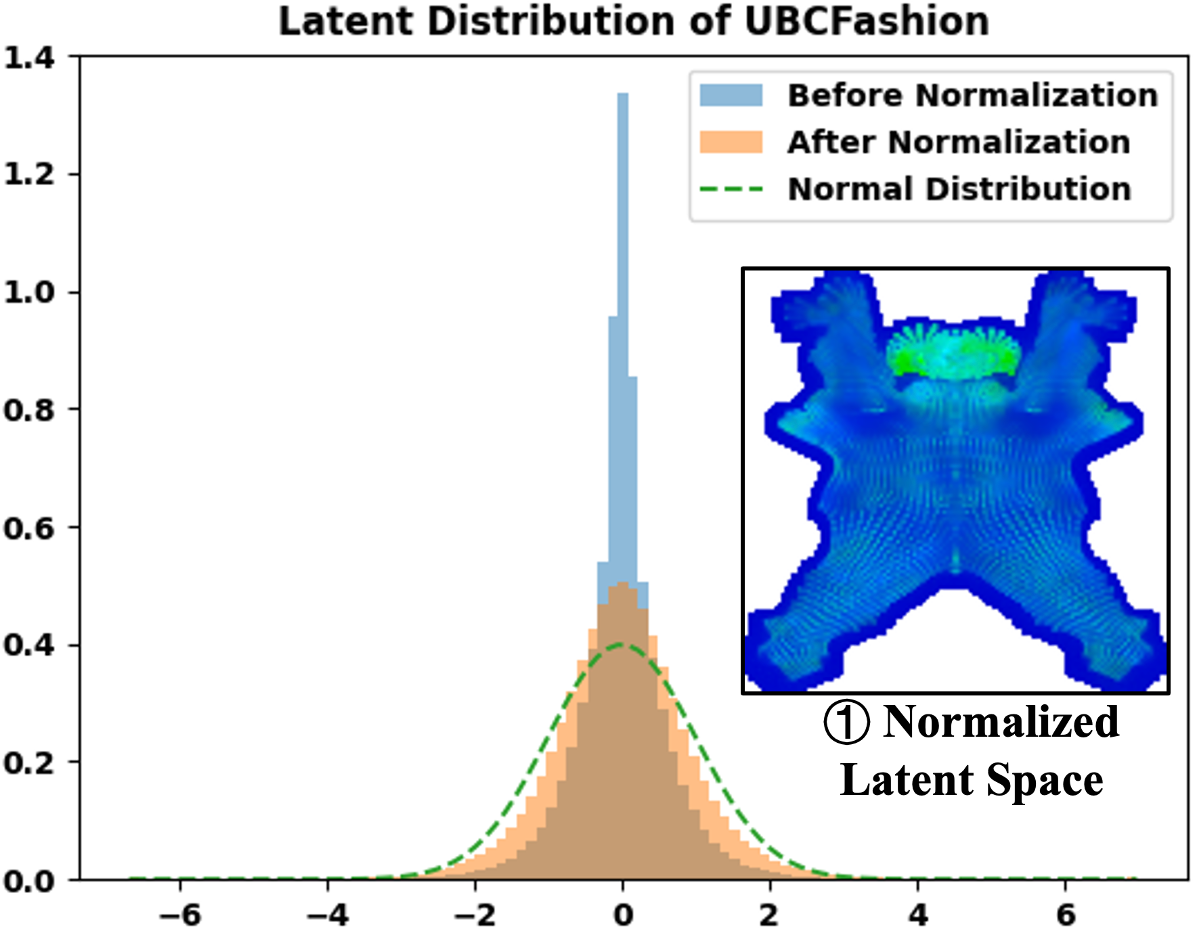}
		\end{center}
        \vspace{-0.1in}
        \captionof{figure}{Ablation of structure-aligned normalization.}
        \vspace{-0.06in}
        \label{fig:ab_diff_norm}
	\end{minipage}
    \begin{minipage}{\linewidth}
		\begin{center} 
			\includegraphics[width=.85\linewidth]{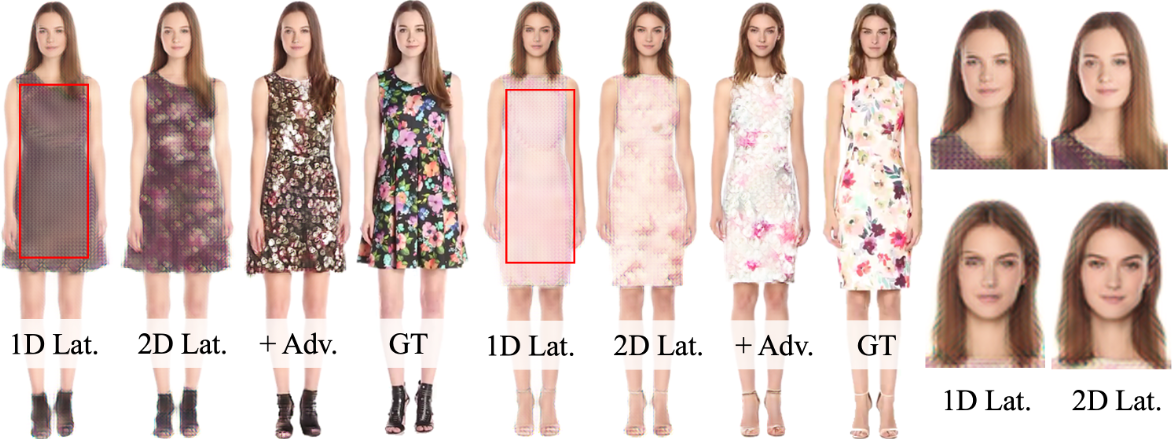}
		\end{center}
        \vspace{-0.06in}
        \captionof{figure}{Ablation study of learning auto-decoder.}
        \label{fig:ab_ad_qual}
	\end{minipage}      
    \vspace{-0.22in}
\end{table*}
}

\newcommand{\tabAbTab}{
\begin{table*}[t]
    \vspace{-0.1in}
    \begin{minipage}{.45\linewidth}
        \begin{center}  
            \includegraphics[width=\linewidth]{img/ab_latent.png}
        \end{center}            
        \vspace{-0.1in}
        \captionof{figure}{Ablation of latent size in auto-decoder.}
        \vspace{-0.06in}
        \label{fig:ab_ad_quan_latent}
    \end{minipage} \hfill
	\begin{minipage}{.5\linewidth}
		\begin{center} 
			\includegraphics[width=\linewidth]{img/ab_normalization_2.png}
		\end{center}
        \vspace{-0.1in}
        \captionof{figure}{Ablation of structure-aligned normalization.}
        \vspace{-0.06in}
        \label{fig:ab_diff_norm}
	\end{minipage}
    \vspace{-0.16in}
\end{table*}
}

\newcommand{\tabQt}{
\vspace{-0.2pt}
\begin{table*}[th]    
	\begin{minipage}{.58\linewidth}\small
		\caption{Quantitative comparisons with SOTA methods on FID@50K $\downarrow$. For reference, we report the quantitative results from the EVA3D {\scriptsize{[ICLR'23]}}, AG3D {\scriptsize{[ICCV'23]}} and PrimDiff {\scriptsize [NeurIPS'23]} paper. }
    \vspace{-0.12pt}
    \begin{tabular}{lccc}
        \specialrule{.1em}{.1em}{.1em}	
        Methods  & {\scriptsize UBCFashion} & {\scriptsize RenderPeople} & {\scriptsize THUman2.0} \\ \hline
        StyleSDF~\cite{stylesdf} & 18.52      & 51.27        &     -      \\ 
        EG3D~\cite{eg3d}     & 23.95      & 24.32        &     -      \\ 
        EVA3D~\cite{eva}    & 12.61      & 44.37        &  {124.54}       \\ 
        AG3D~\cite{ag3d}     & 11.04      &  -           &     -      \\ 
        PrimDiff~\cite{chen2023primdiffusion} & -  &  17.95           & -  \\ \hline        
        Ours     & \textbf{9.56}     & \textbf{13.98}         & \textbf{25.22}     \\ 
            \specialrule{.1em}{.1em}{.1em}	

    \end{tabular}		
    \vspace{-0.12in}
    \label{tab:cmpquant}
	\end{minipage} \hfill      
    \begin{minipage}{.38\linewidth}\small
		\begin{center}
              \captionof{table}{Ablation of normalization in diffusion on FID@4K on UBCFashion. Part-aware normalization normalizes each body part latent locally by segmentations (see Fig. \ref{fig:partediting}).}
              \vspace{-0.2pt}
			\begin{tabular}{l|l}
                \specialrule{.1em}{.1em}{.1em}	

                    Norm. Method & FID $\downarrow$     \\ \hline
                    None                   & 20.13 \\ 
                    Standard         & 24.86 \\ 
                    Part-aware             & 25.02 \\ 
                    Struct-aligned & 19.58 \\ 
                        \specialrule{.1em}{.1em}{.1em}	

                \end{tabular}
                \label{tab:ab_diff_quant}
                \vspace{-0.12in}               
            \end{center}
	\end{minipage}
  \vspace{-0.1in}               
\end{table*}
}

\newcommand{\tabQtDF}{
\begin{table}[]
	\renewcommand{\arraystretch}{1.3}
	\centering
	\caption{Quantitative results on DeepFashion. For reference, we report the quantitative results from the EVA3D (marked by `*') and the AG3D paper (marked by `$\circ$').}
    \vspace{-0.12pt}
    \begin{tabular}{lc}
       \hline
       
        FID $\downarrow$  & DeepFashion \\ \hline
        StyleSDF$^*$ ~\cite{stylesdf} & 92.40  \\ 
        EG3D$^*$~\cite{eg3d}     & 26.38      \\ 
        ENARF-GAN$^*$~\cite{enarfgan}     & 77.03      \\ 
        EVA3D$^*$~\cite{eva}    & 15.91      \\    EVA3D(Public)$^{\circ}$~\cite{eva}    & 20.45      \\        
        AG3D$^{\circ}$~\cite{ag3d}     & 10.93     \\ \hline
        Ours     & {20.82}     \\ \hline
    \end{tabular}		
    \vspace{-0.12in}
    \label{tab:quant_df}
    \end{table}
}

\newcommand{\figUsCmpPd}{
\begin{table*}[t]
    \begin{minipage}{.48\linewidth}
        \begin{center}
					\includegraphics[width=\linewidth]{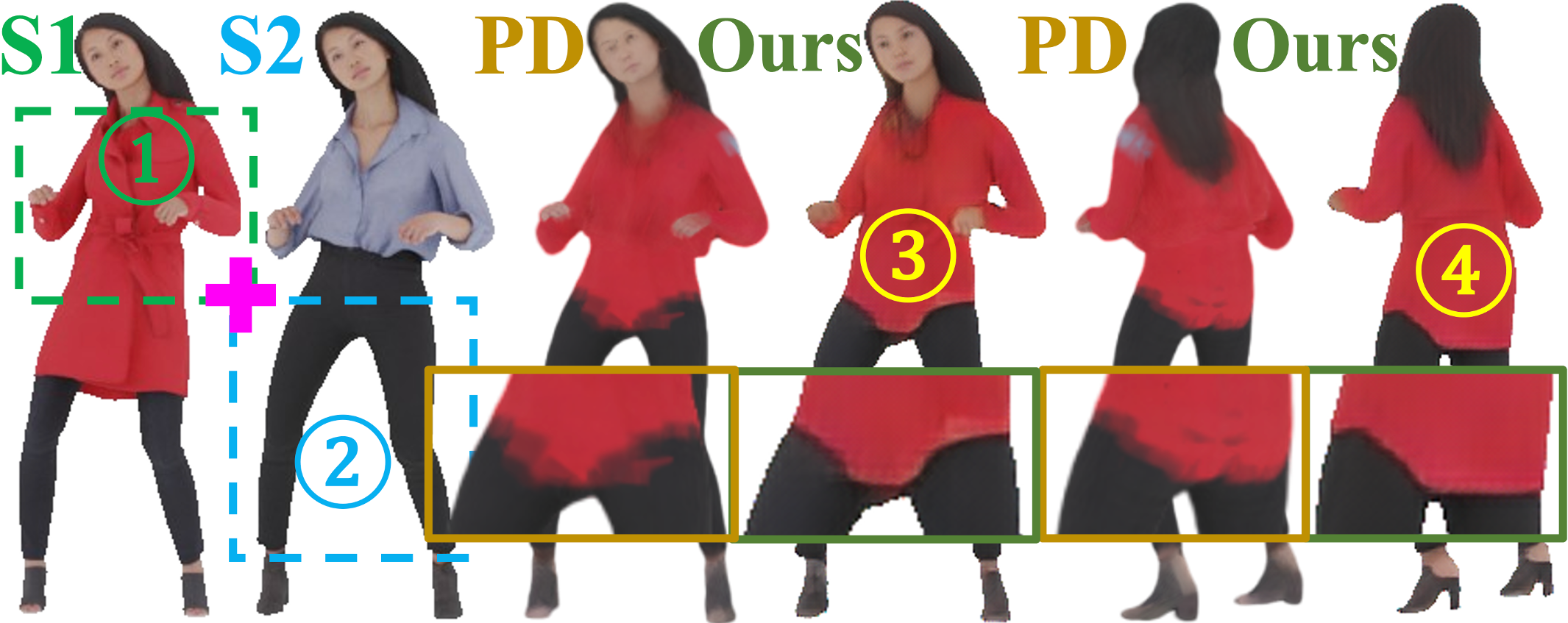}
        \end{center}
        \vspace{-0.1in}
        \captionof{figure}{Comparisons with PrimDiff (PD) on texture transfer. Ours achieves better results \cnum{3}\cnum{4} by mixing the source \cnum{1}\cnum{2}.}
    \label{fig:cmpPd}        
    \end{minipage} \hfill
	\begin{minipage}{.49\linewidth}
        \begin{center}
					\includegraphics[width=\linewidth]{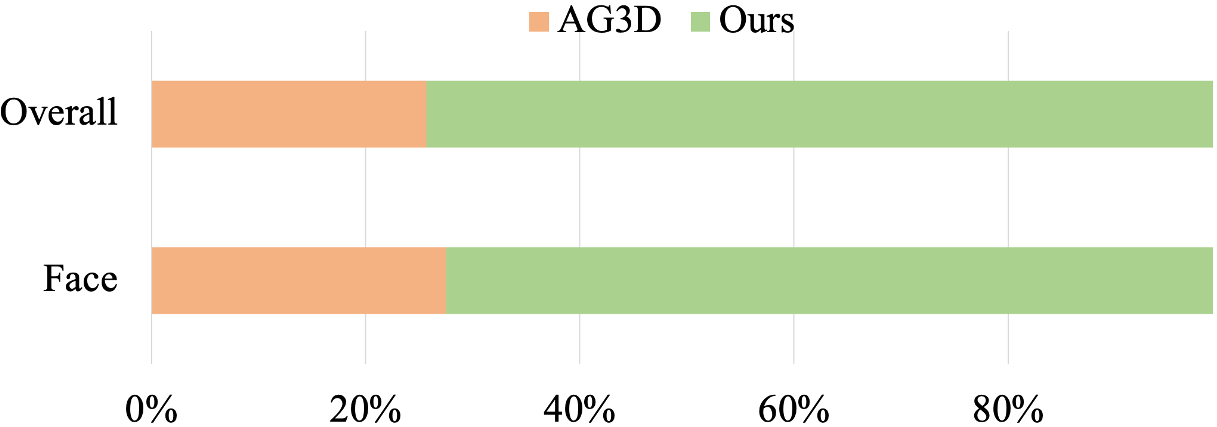}
        \end{center}
        \vspace{-0.1in}
        \captionof{figure}{User study. We report how often the generated images by ours are preferred over AG3D in terms of overall appearance and face quality.}
        \label{fig:userstudy}
    \end{minipage}      
	\vspace{-0.16in}
\end{table*}
}

\newcommand{\figPartInv}{
\begin{table*}[t]
    \begin{minipage}{.71\linewidth}
			\begin{center}
				\includegraphics[width=\linewidth]{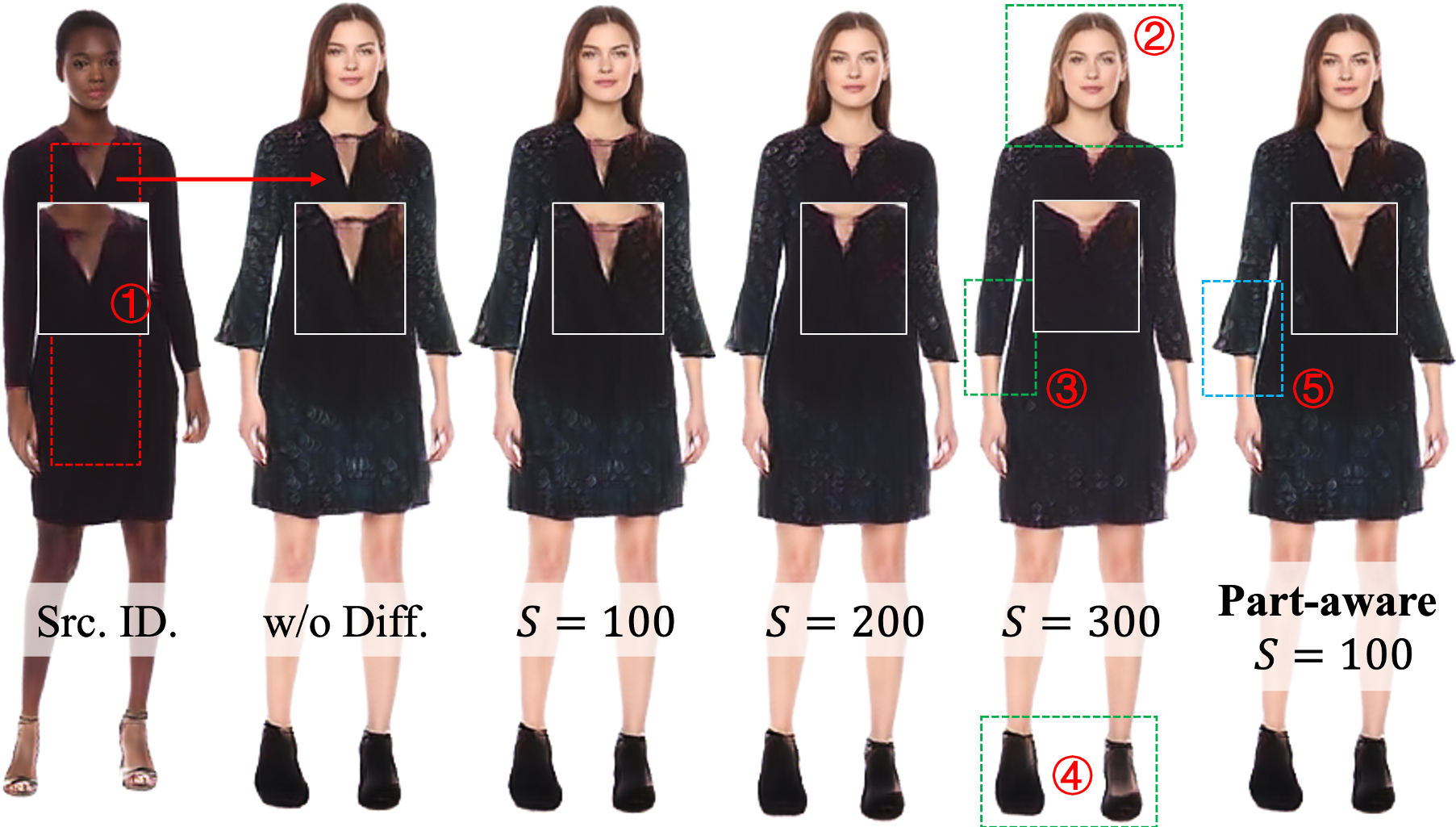}
			\end{center}
			\vspace{-0.12in}
			\captionof{figure}{{Part-aware diffusion (supp.) for local enhancement.}}
			\label{fig:part_diffusion}
    \end{minipage} \hfill
	\begin{minipage}{.27\linewidth}
		\begin{center}
			\includegraphics[width=\linewidth]{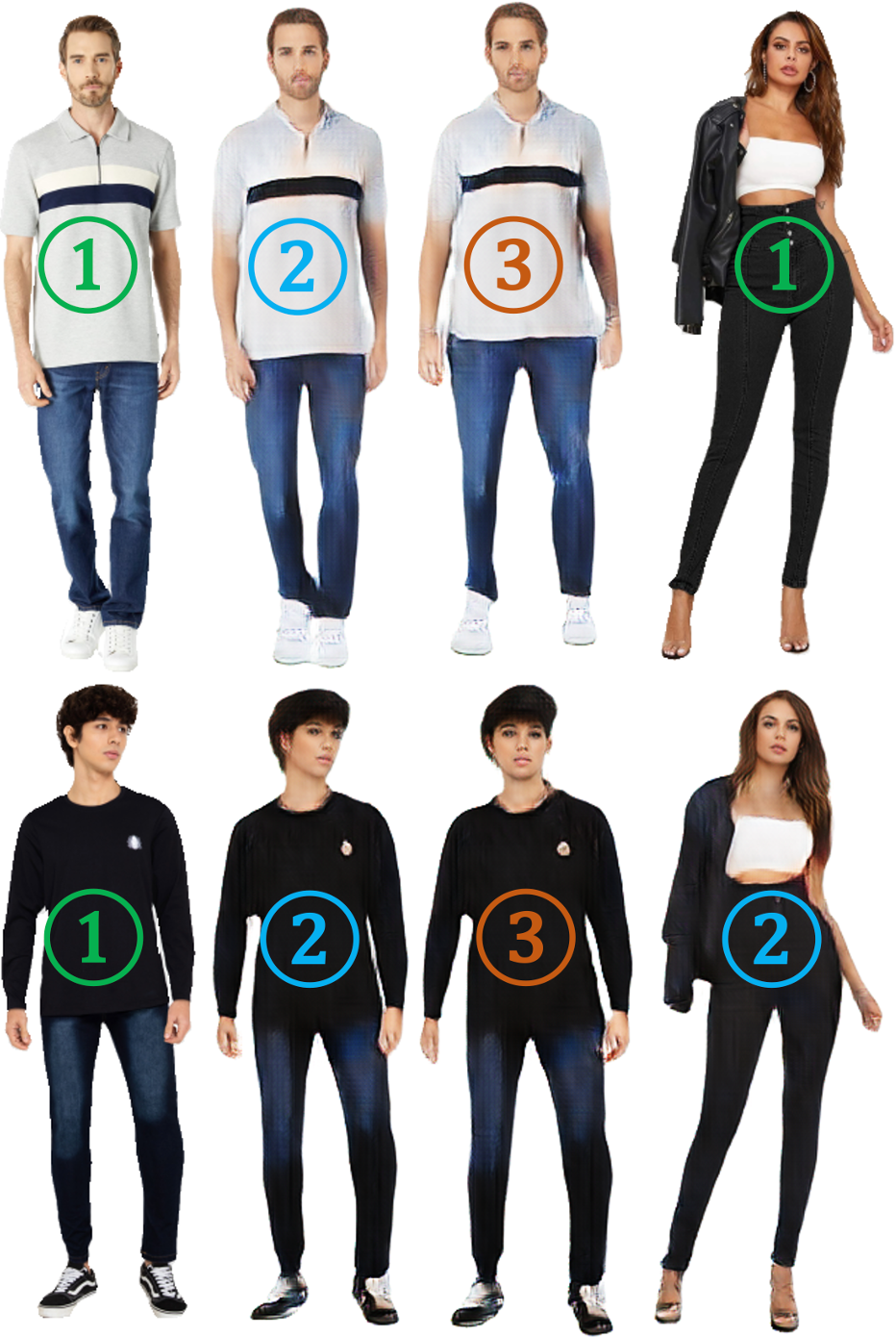}
        \end{center}
        \vspace{-0.14in}
        \captionof{figure}{{Inversion.}}
    \label{fig:appInv}			
  \end{minipage}      
	\vspace{-0.28in}
\end{table*}
}

\newcommand{\figPartRef}{
\begin{figure}[]
	\begin{center}
        \includegraphics[width=.8\linewidth]{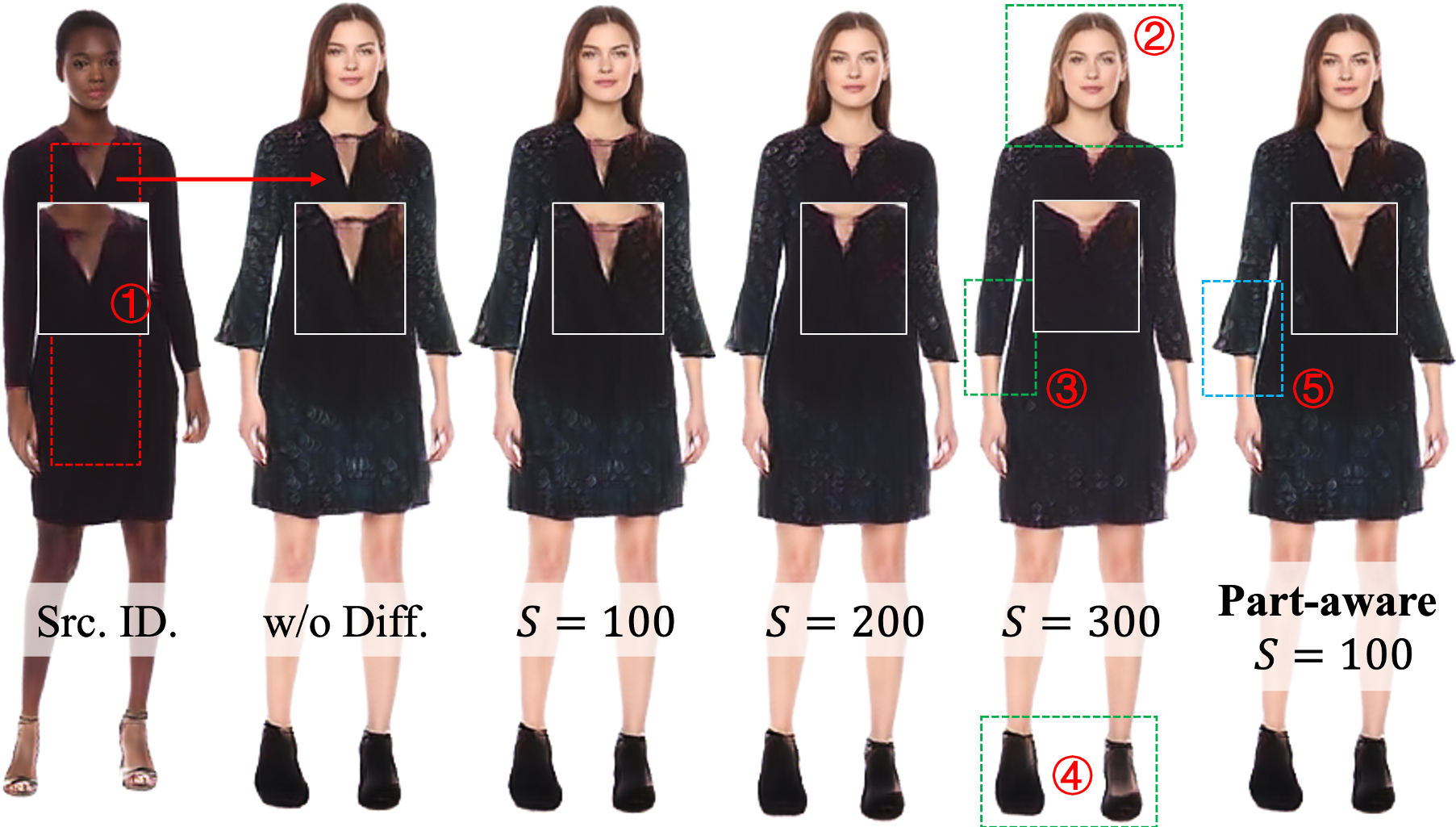}
	\end{center}
	\vspace{-0.12in}
	\caption{Part-aware diffusion for generation enhancement.}
	\label{fig:part_diffusion}
	\vspace{-0.12in}
\end{figure}
}

\newcommand{\figFramework}{
\begin{figure*}[t]
	\begin{center}
		\includegraphics[width=\linewidth]{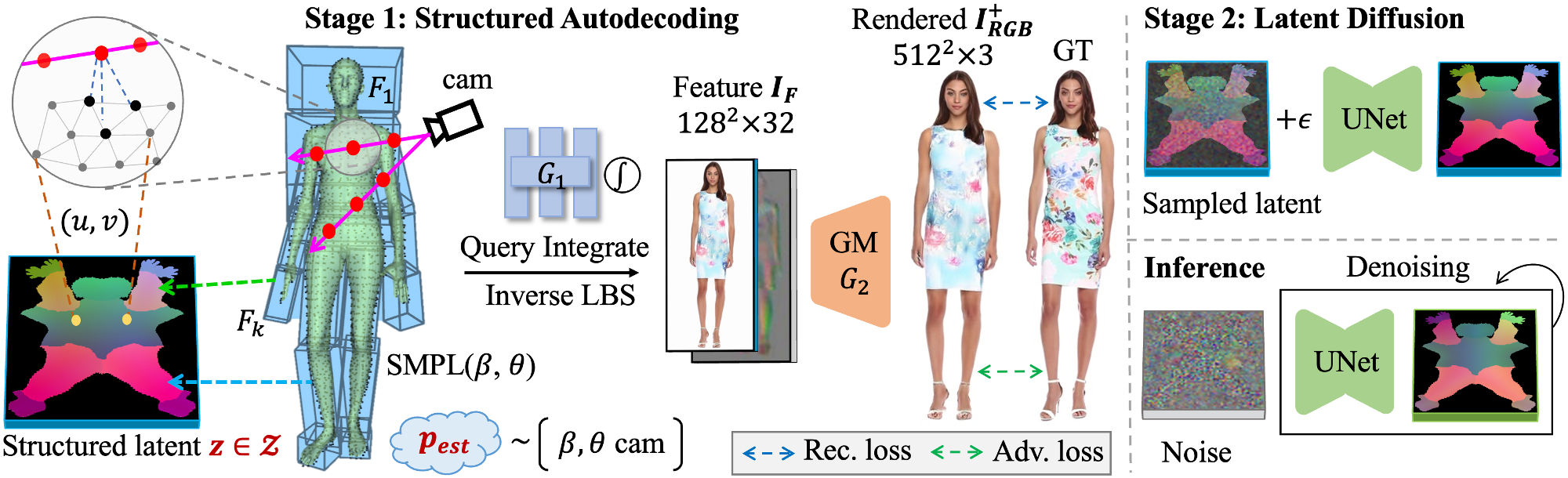}
	\end{center}
	\vspace{-0.16in}
	\caption{Two-stage framework. In Stage 1, given a training dataset containing various human subject images with estimated SMPL and camera parameters distribution $p_{est}$, an auto-decoder is learned to optimize the structured latent $z \in \mathcal{Z}$ for each training subject. Each latent is rendered into a pose- and view-dependent image by a structured volumetric renderer $G_1$ and a global style mixer module (GM) $G_2$. In Stage 2, the auto-decoder parameters are frozen and the learned structured latent $\mathcal{Z}$ are then used to train the latent diffusion model. At inference time, latents are randomly sampled and decoded by $G_2 \circ G_1$ for human rendering.}
	\label{fig:framework}
	\vspace{-0.16in}
\end{figure*}
}

\newcommand{\figSupStyleMix}{
\begin{figure*}[t]
	\begin{center}
        \includegraphics[width=\linewidth]{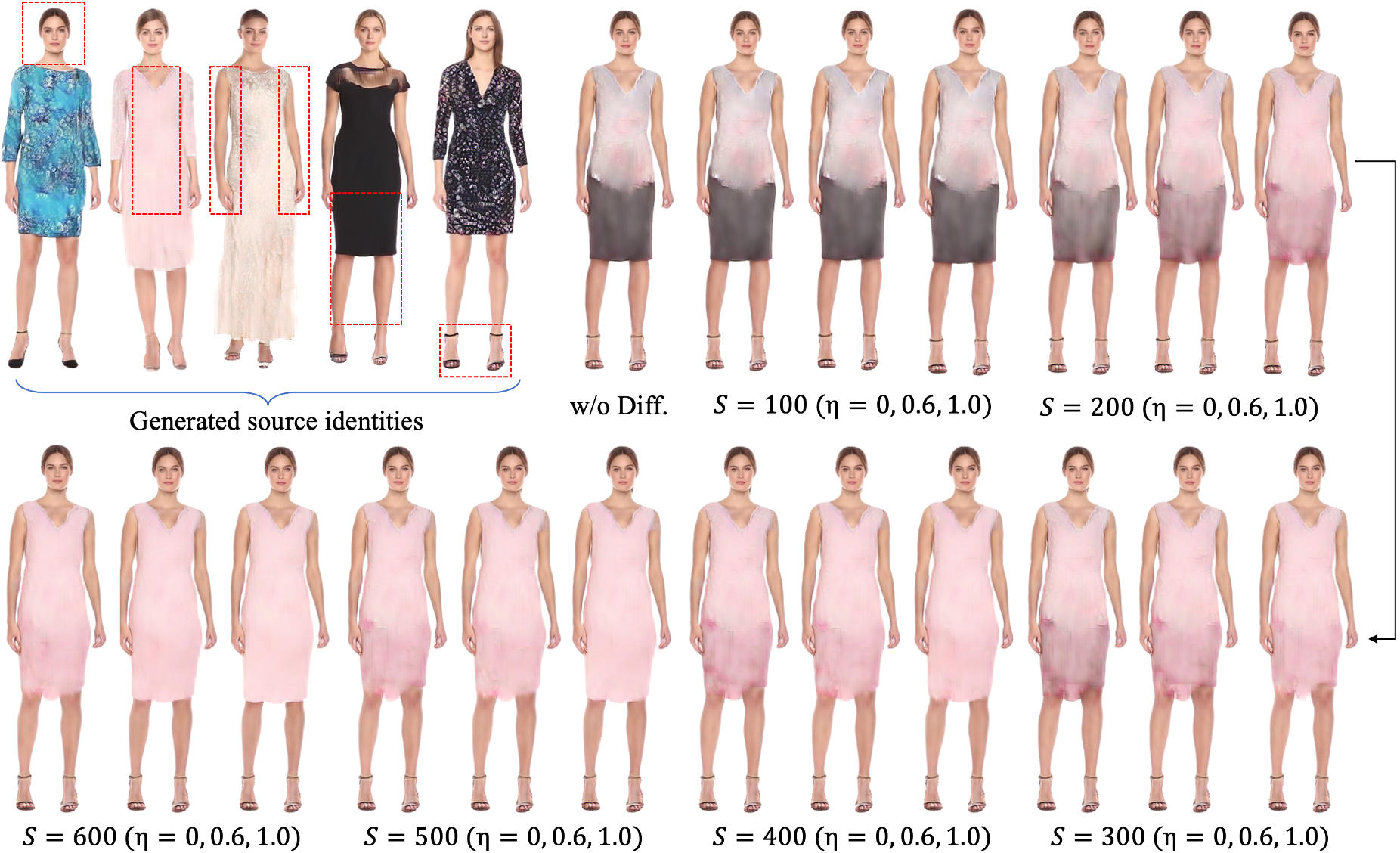}
	\end{center}
 	\vspace{-0.14in}
	\caption{The effect of latent diffusion in compositional generations.}
	\label{fig:sup_stylemixing}
	\vspace{-0.16in}  
\end{figure*} 
}        

\newcommand{\figCtrlPart}{
\begin{table*}[t]
	\vspace{-0.06in}
	\begin{minipage}{\linewidth}
		\begin{center}
			\includegraphics[width=\linewidth]{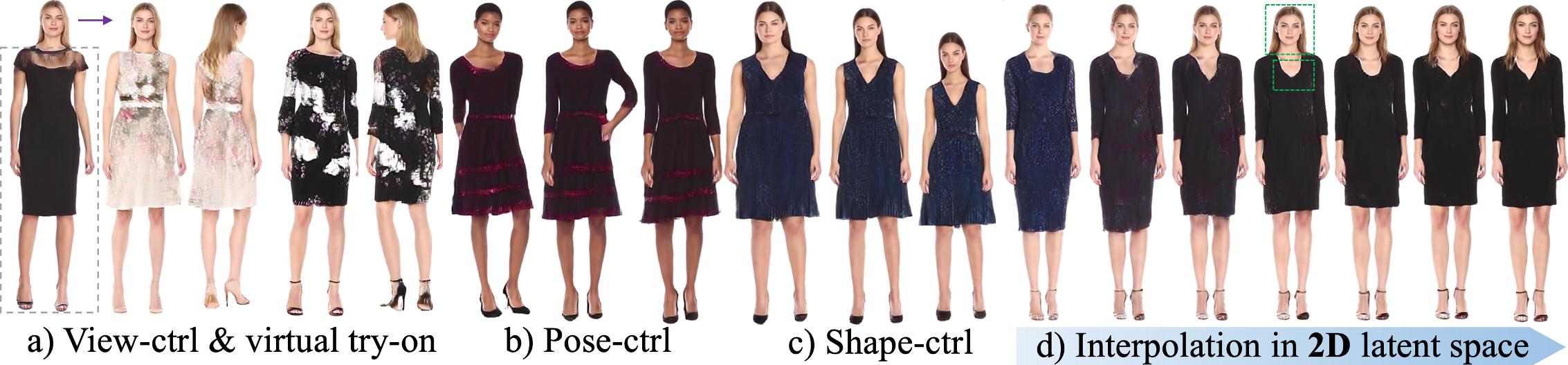}
		\end{center}
		\vspace{-0.08in}
		\captionof{figure}{Controllable generations. The 3D-aware architecture with inherent human body priors enables explicit control over rendering views, human poses, and shapes. We can also smoothly interpolate two samples on the 2D latent space in a similar way to the interpolation on the 1D latent space.}
		\label{fig:ctrl}
		\vspace{-0.10in}
	\end{minipage} 
\begin{minipage}{\linewidth}
	\begin{center}
		\includegraphics[width=\linewidth]{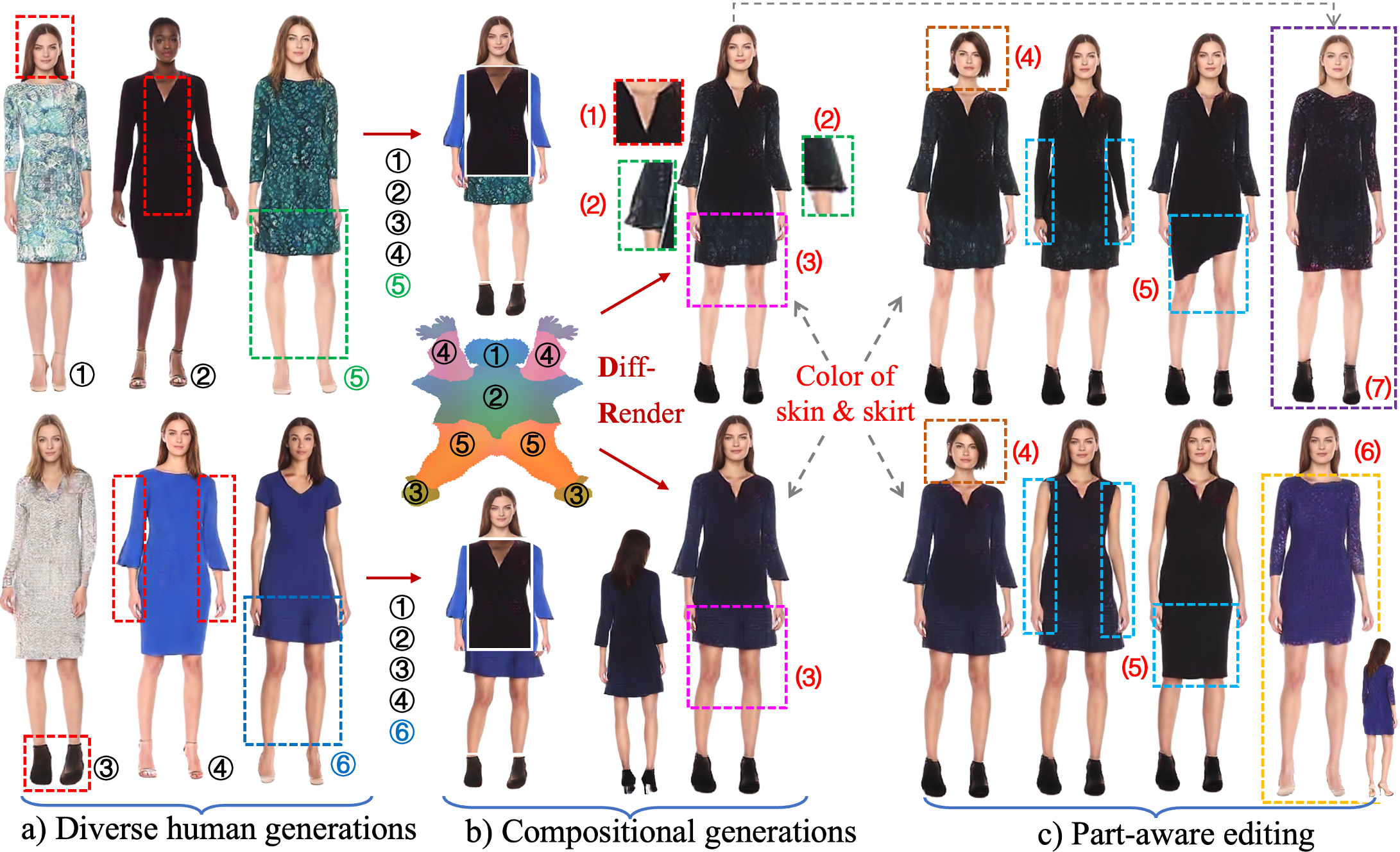}
	\end{center}
	\vspace{-0.1in}
	\captionof{figure}{\nickname{} enables compositional 3D human generation and part-aware editing. Taking six body parts from a), coherent composition and blending results can be achieved in b). Using the Diff-Render procedure, part-aware editing enables lots of downstream tasks in c).
 }
	\label{fig:partediting}
	\vspace{-0.24in}
	\end{minipage}      
\end{table*}
}

\newcommand{\figPartEditing}{
\begin{figure*}[]
	\begin{center}
		\includegraphics[width=\linewidth]{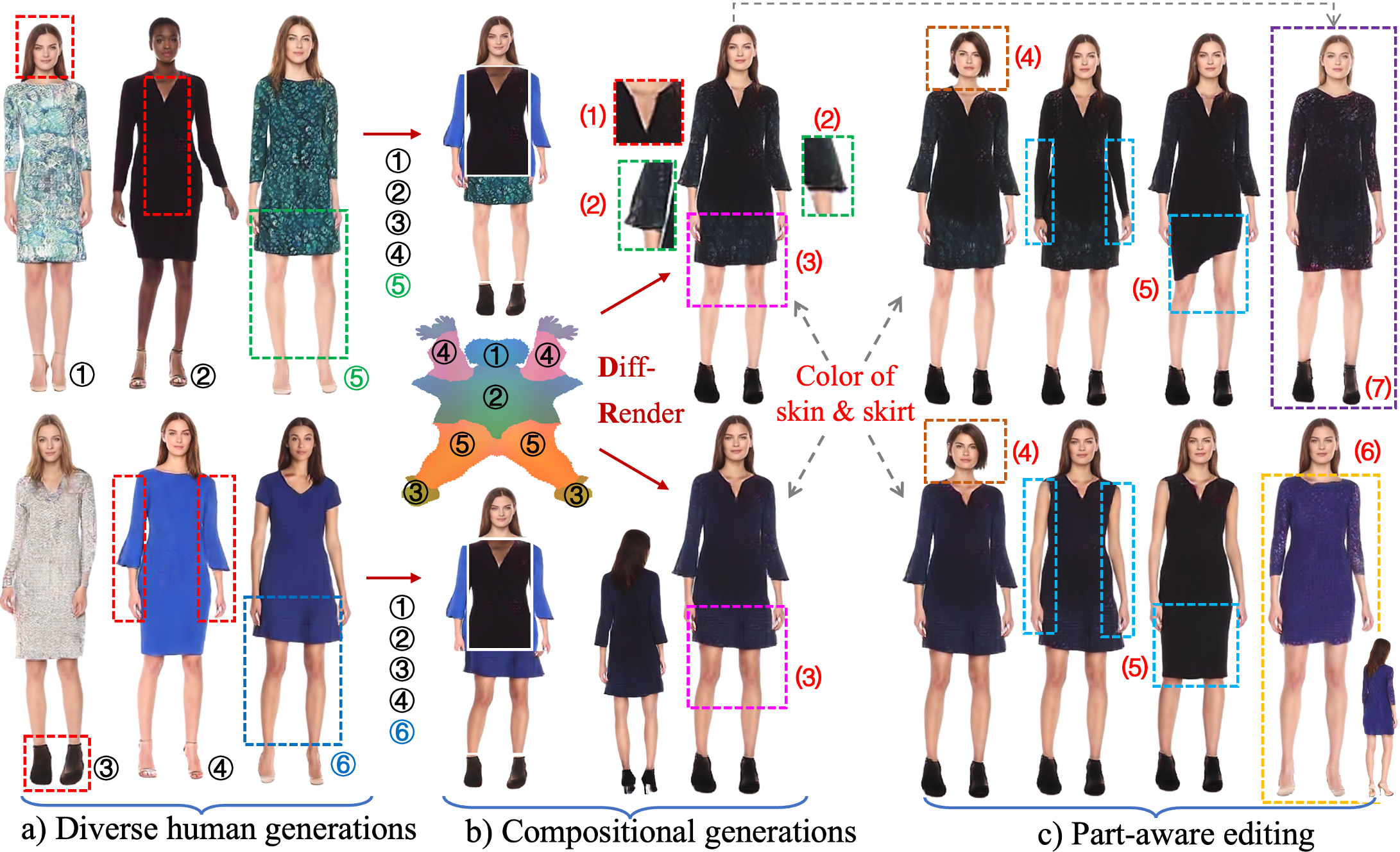}
	\end{center}
	\vspace{-0.16in}
	\caption{\nickname{} enables compositional 3D human generation and part-aware editing. Taking six body parts from a), coherent composition and blending results can be achieved in b). Using the Diff-Render procedure, part-aware editing enables lots of downstream tasks in c).
 }
	\label{fig:partediting}
	\vspace{-0.16in}
\end{figure*}
}

\newcommand{\figCtrl}{
\begin{figure*}[t]
	\begin{center}
        \includegraphics[width=\linewidth]{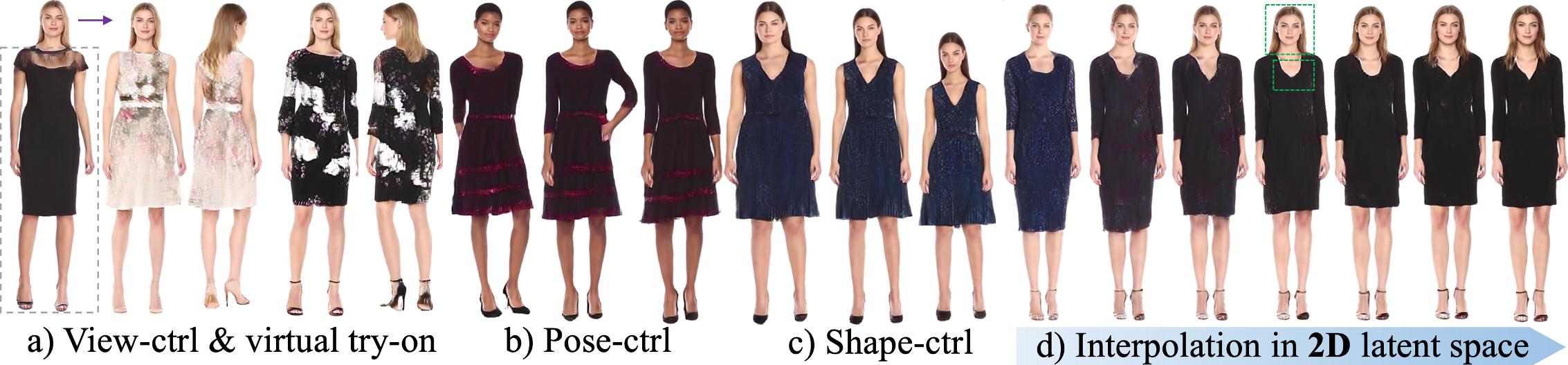}
	\end{center}
	\vspace{-10pt}
	\caption{Controllable generations. The 3D-aware architecture with inherent human body priors enables explicit control over rendering views, human poses, and shapes. We can also smoothly interpolate two samples on the 2D latent space in a similar way to the interpolation on the 1D latent space.}
	\label{fig:ctrl}
	\vspace{-0.16in}
\end{figure*}
}

\newcommand{\figUserStudy}{
\begin{figure}[]
	\begin{center}
        \includegraphics[width=\linewidth]{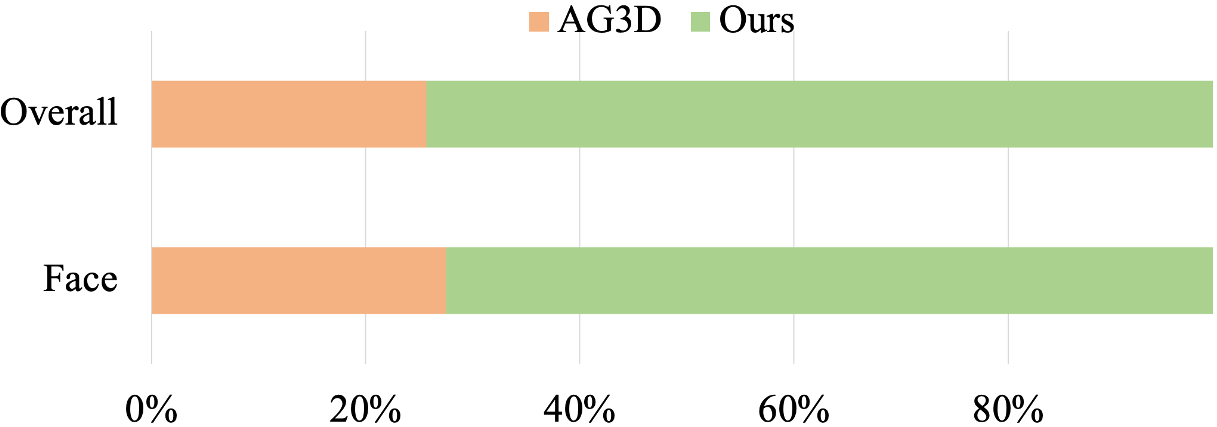}
	\end{center}
	\vspace{-0.12in}
	\caption{User study. We conduct a perceptual study and report how often the generated images by our method are preferred over AG3D in terms of overall appearance quality and face quality.}
	\label{fig:userstudy}
	\vspace{-0.12in}
\end{figure}
}

\newcommand{\figAbAd}{
\begin{figure}[t]
	\begin{center}
        \includegraphics[width=.8\linewidth]{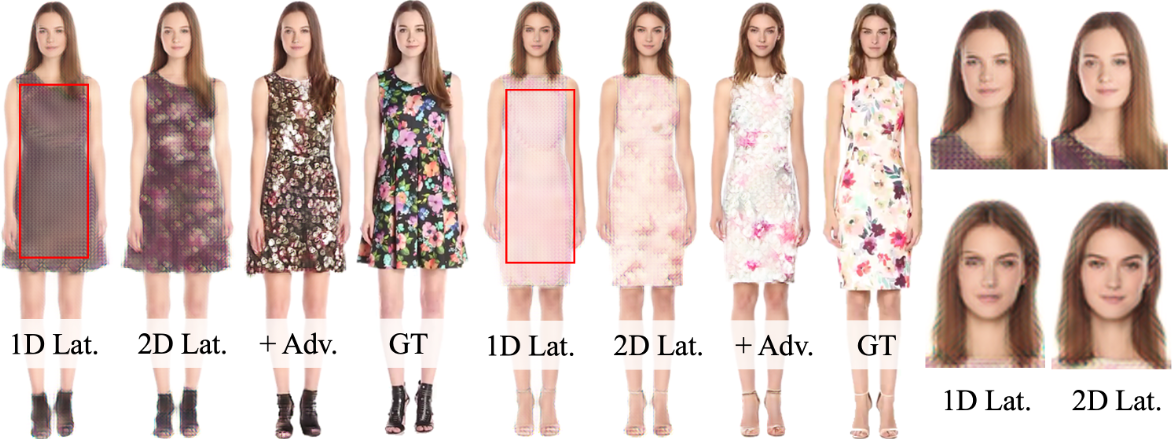}
	\end{center}
	\vspace{-0.16in}
	\caption{Ablation study of learning auto-decoder.}	\label{fig:ab_ad_qual}
	\vspace{-0.22in}
\end{figure}
}

\newcommand{\figAbadInv}{
\begin{table*}[t]
    \begin{minipage}{.78\linewidth}
			\begin{center}
        \includegraphics[width=\linewidth]{img/ab_lat12D_face.png}
		\end{center}
		\vspace{-0.12in}
		\captionof{figure}{Ablation study of learning auto-decoder.}	\label{fig:ab_ad_qual}
		\vspace{-0.16in}
  \end{minipage} \hfill
	\begin{minipage}{.20\linewidth}
        \begin{center}
    			\includegraphics[width=\linewidth]{img/inversion.png}
        \end{center}
        \vspace{-0.14in}
        \captionof{figure}{\scriptsize{Inversion}}
    \label{fig:appInv}
    \vspace{-0.16in}
    \end{minipage}      
\end{table*}
}

\newcommand{\figCmpUBC}{
\begin{figure*}[t]
	\begin{center}
        \includegraphics[width=\linewidth]{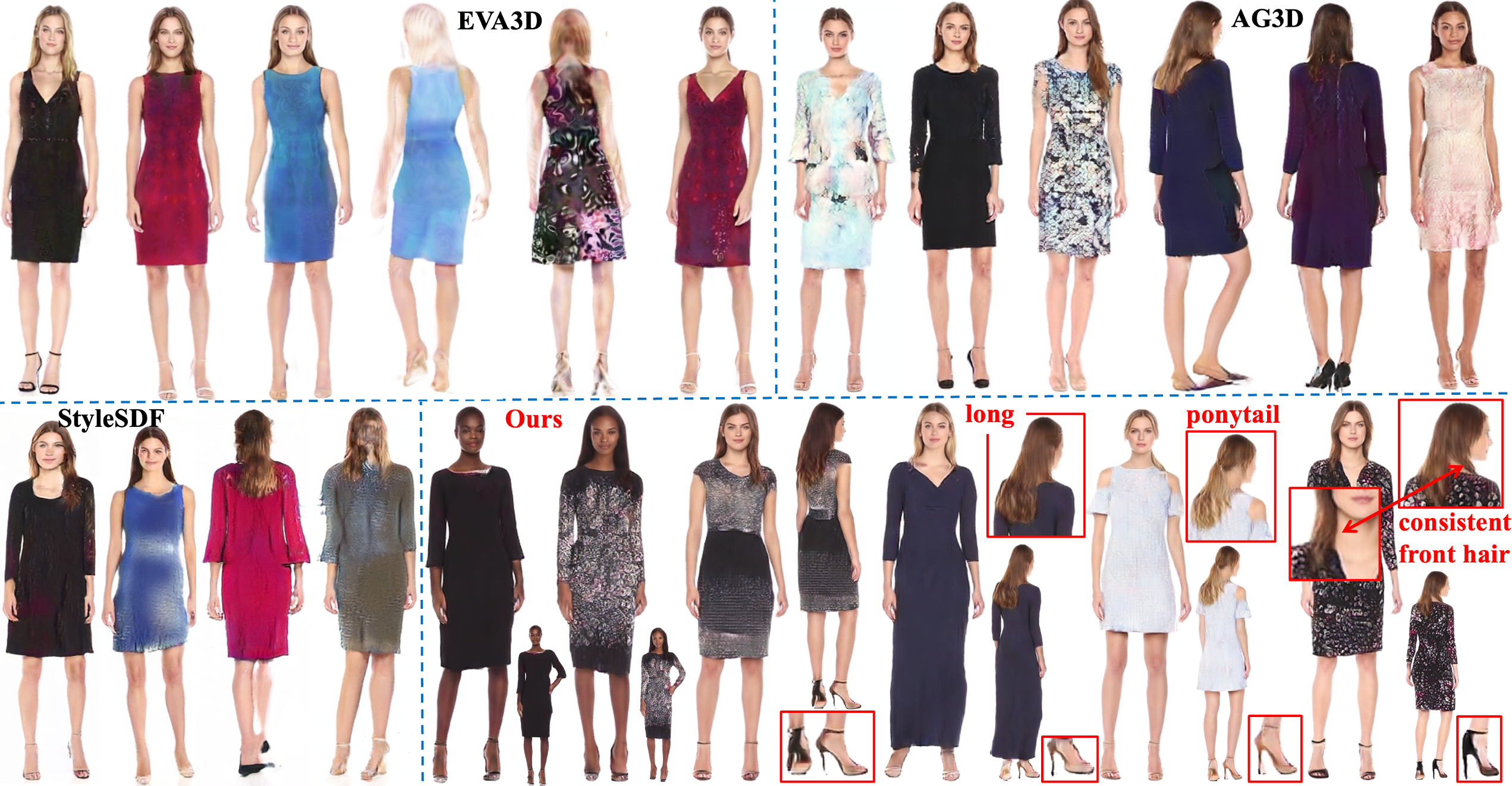}
	\end{center}
	\vspace{-0.16in}
	\caption{Qualitative results on UBCFashion. We generate diverse view-consistent humans under different poses/views for different clothing styles (\eg dress) and hairstyles.}
	\label{fig:cmp_ubc}
	\vspace{-0.18in}
\end{figure*}
}

\newcommand{\figCmpRP}{
\begin{figure*}[h]
	\begin{center}
        \includegraphics[width=\linewidth]{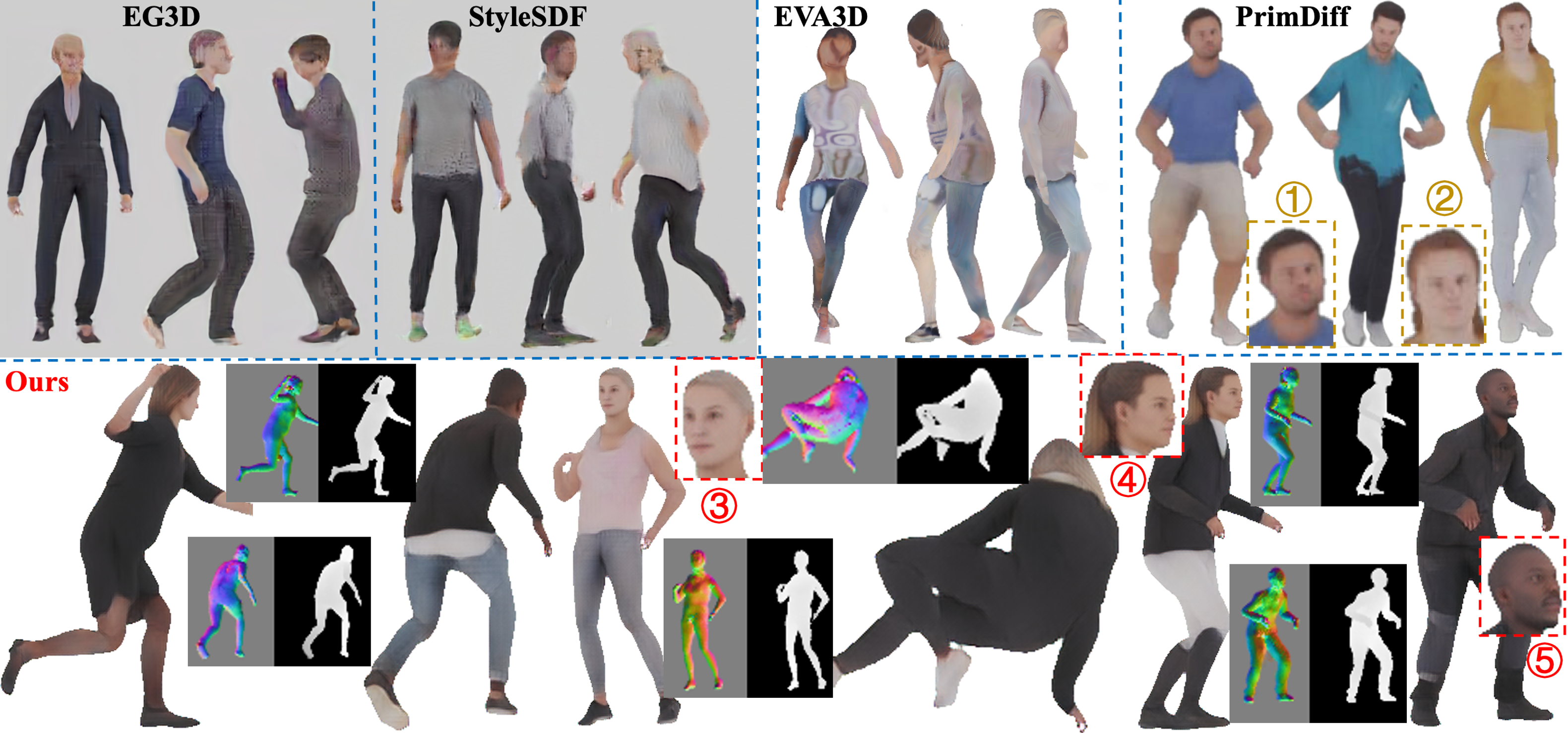}
	\end{center}
	\vspace{-0.18in}
	\caption{Qualitative comparisons on RenderPeople \cite{renderpeople}. The geometry is visualized as  normal/depth maps at $128\times64$ resolution. The rendered images are cropped to $512\times256$ in visualization. We synthesize high-quality faces \cnum{3}\cnum{4}\cnum{5} vs. \cnum{1}\cnum{2} PrimDiff \cite{chen2023primdiffusion}. }
	\label{fig:cmp_rp}
	\vspace{-0.16in}
\end{figure*}
}

\newcommand{\figQtThuRp}{
\begin{table*}[t]
    \begin{minipage}{.45\linewidth}
        \begin{center}
        \includegraphics[width=\linewidth]{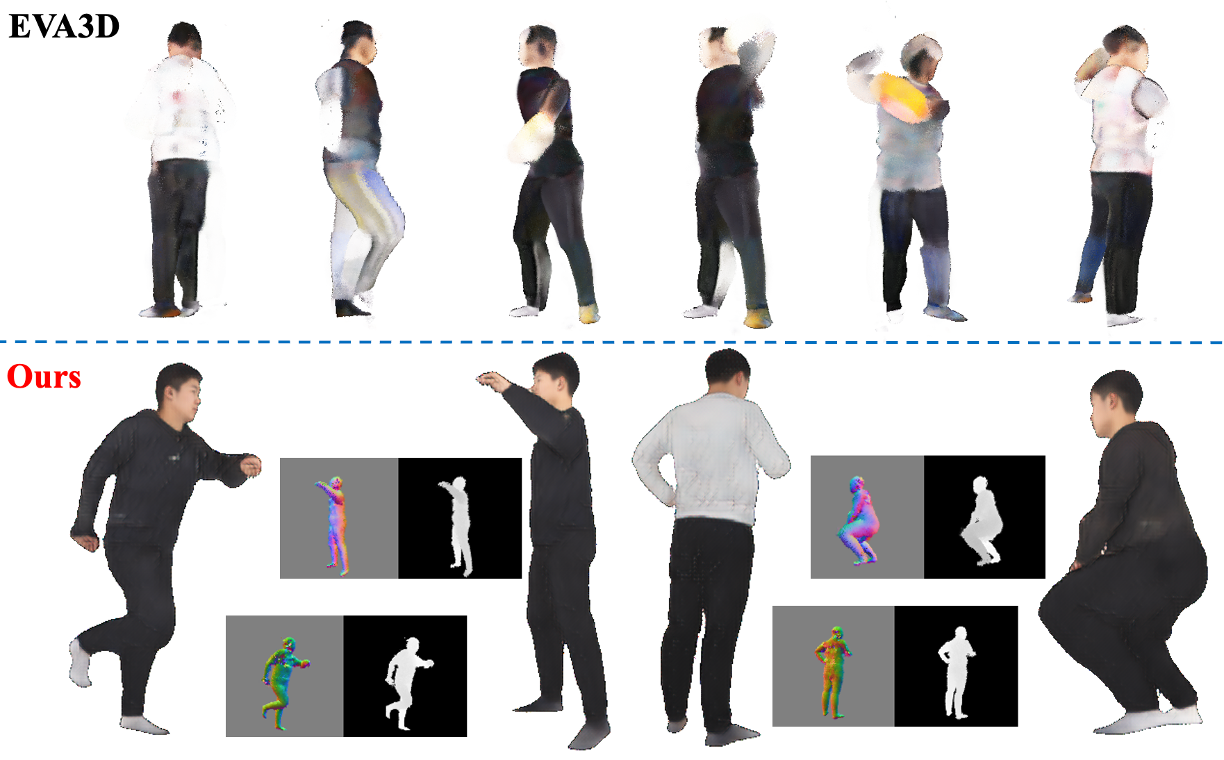}
	\end{center}
 	\vspace{-0.12in}
	\captionof{figure}{Comparisons on THUman2.0 \cite{thuman2}. The geometry is visualized as  normal/depth maps at $128\times64$ resolution.}
	\label{fig:sup_thuman}
	\vspace{-0.2in}
    \end{minipage} \hfill
	\begin{minipage}{.53\linewidth}
       \begin{center}
        \includegraphics[width=\linewidth]{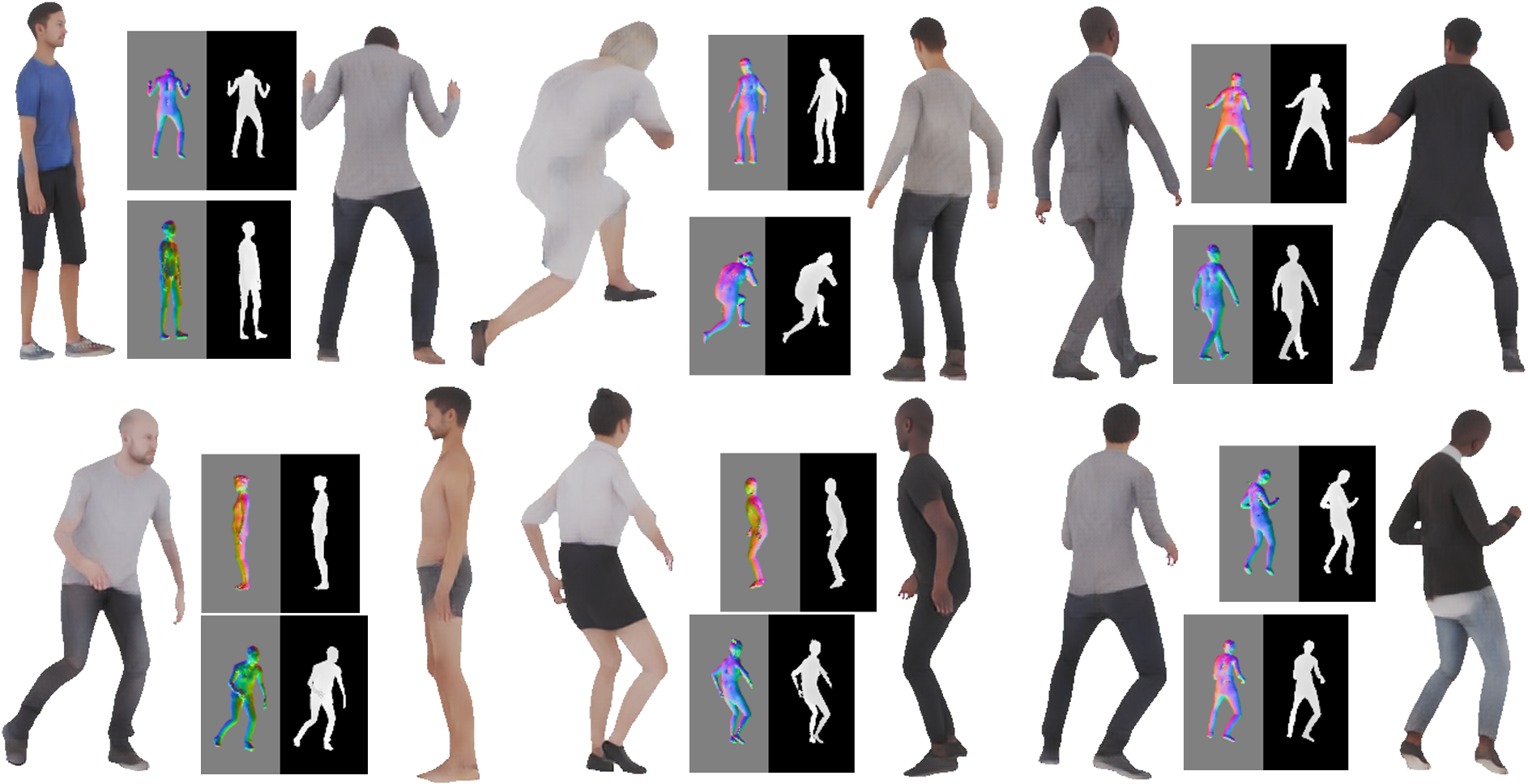}
	\end{center}
 	\vspace{-0.12in}
	\captionof{figure}{Qualitative results on RenderPeople. The geometry is visualized as  normal/depth maps at $128\times64$ resolution.}
	\label{fig:sup_renderpeople}
	\vspace{-0.2in}
    \end{minipage}      
\end{table*}
\vspace{-0.2in}
}

\newcommand{\figQtThuDf}{
\begin{table*}[t]
	\begin{minipage}{\linewidth}
		\begin{center}
			\includegraphics[width=\linewidth]{img/cmp_rp2.png}
		\end{center}
		\vspace{-0.12in}
		\captionof{figure}{Qualitative comparisons on RenderPeople \cite{renderpeople}. The geometry is visualized as  normal/depth maps at $128\times64$ resolution, and images are cropped to $512\times256$ for visualization. We synthesize high-quality faces \cnum{3}\cnum{4}\cnum{5} vs. \cnum{1}\cnum{2} PrimDiff \cite{chen2023primdiffusion}. }
		\label{fig:cmp_rp}
		\vspace{-0.1in}
	\end{minipage} 
  \begin{minipage}{.66\linewidth}
        \begin{center}
        \includegraphics[width=\linewidth]{img/cmp_thuman.png}
	\end{center}
 	\vspace{-0.12in}
	\captionof{figure}{Comparisons on THUman2.0 \cite{thuman2}. The geometry is visualized as  normal/depth maps at $128\times64$ resolution.}
	\label{fig:sup_thuman}
	\vspace{-0.2in}
    \end{minipage} \hfill
	\begin{minipage}{.32\linewidth}
       \begin{center}
        \includegraphics[width=\linewidth]{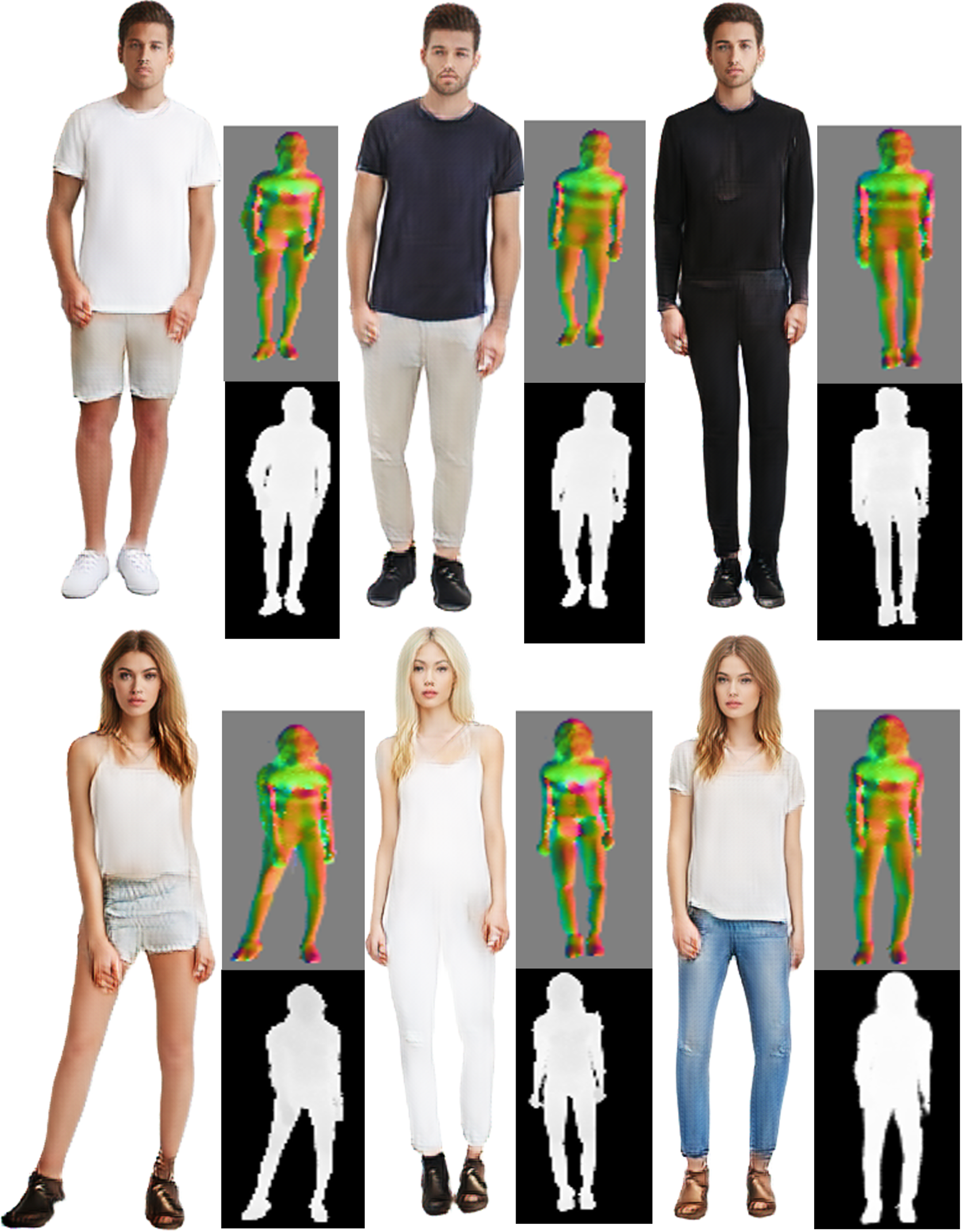}
	\end{center}
 	\vspace{-0.12in}
	\captionof{figure}{Qualitative generations on DeepFashion \cite{Liu2016DeepFashion}.}
	\label{fig:df}
	\vspace{-0.26in}
    \end{minipage}      
\end{table*}
}
\newcommand{\figCmpThuman}{
\begin{figure}[]
	\begin{center}
        \includegraphics[width=\linewidth]{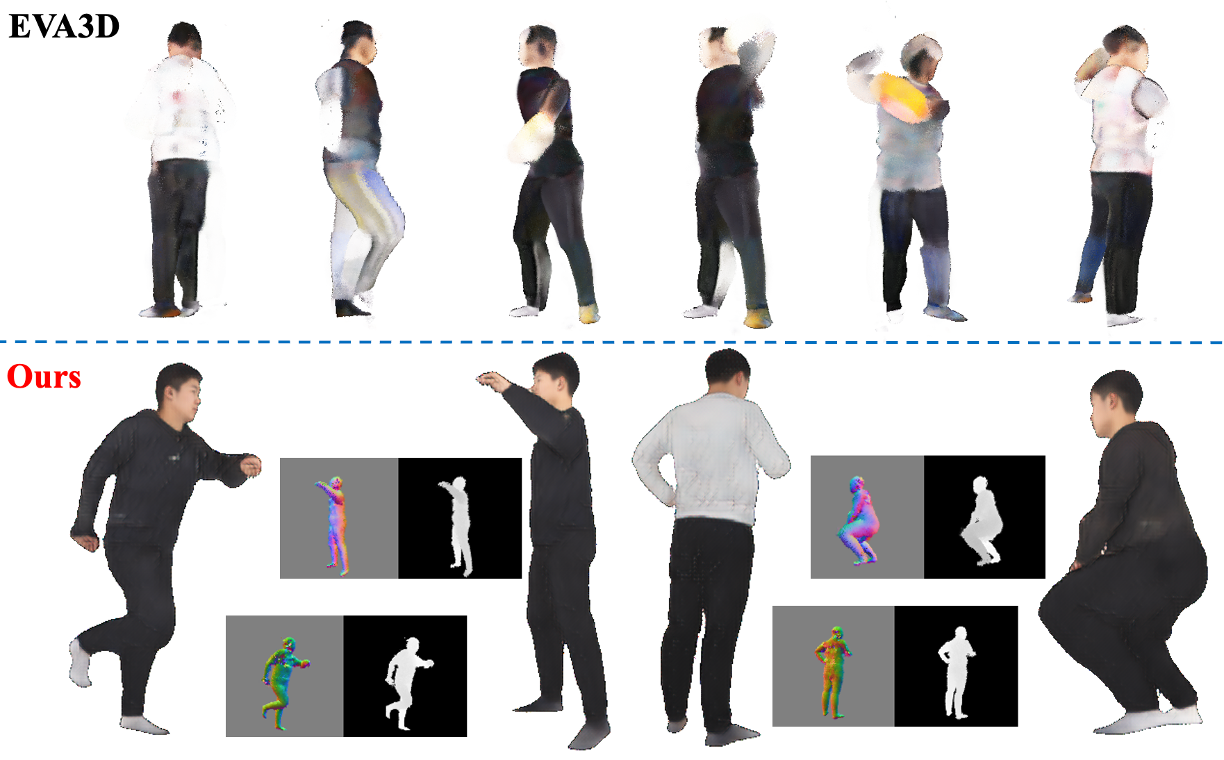}
	\end{center}
 	\vspace{-0.06in}
	\caption{Qualitative comparisons on THUman2.0 \cite{thuman2}. For our method, normal and depth maps at $128\times64$ resolution are rendered by our volumetric renderer.}
	\label{fig:sup_thuman}
	\vspace{-0.12in}
\end{figure}
}

\newcommand{\figDfVis}{
\begin{figure}[t]
	\begin{center}
        \includegraphics[width=\linewidth]{img/deepfashion_vis.png}
	\end{center}
 	\vspace{-0.06in}
	\caption{Generated results on DeepFashion \cite{Liu2016DeepFashion}.}
	\label{fig:sup_df}
	\vspace{-0.12in}
\end{figure}
}

\newcommand{\figEditWild}{
\begin{figure}[t]
	\begin{center}
        \includegraphics[width=.9\linewidth]{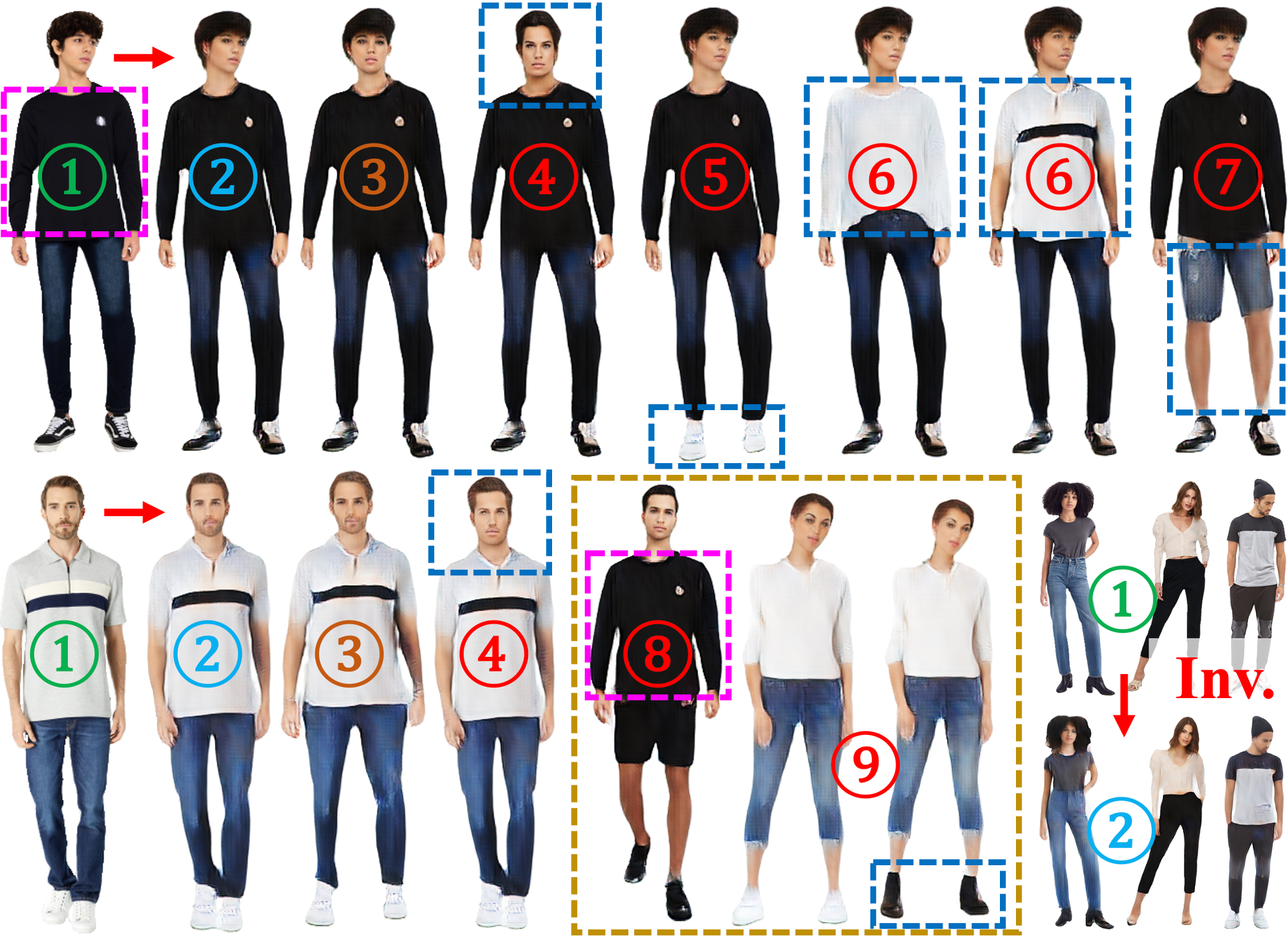}
	\end{center}
 	\vspace{-0.06in} 
	\caption{Editing in-the-wild images. To edit in-the-wild images \cnum{1}, we first apply  inversion \cnum{2}, and edit the images via part-aware diffusion.}
	\label{fig:sup_edit_wild}
	\vspace{-0.12in}
\end{figure}
}

\newcommand{\figDfLatent}{
\begin{figure}[t]
	\begin{center}        \includegraphics[width=.7\linewidth]{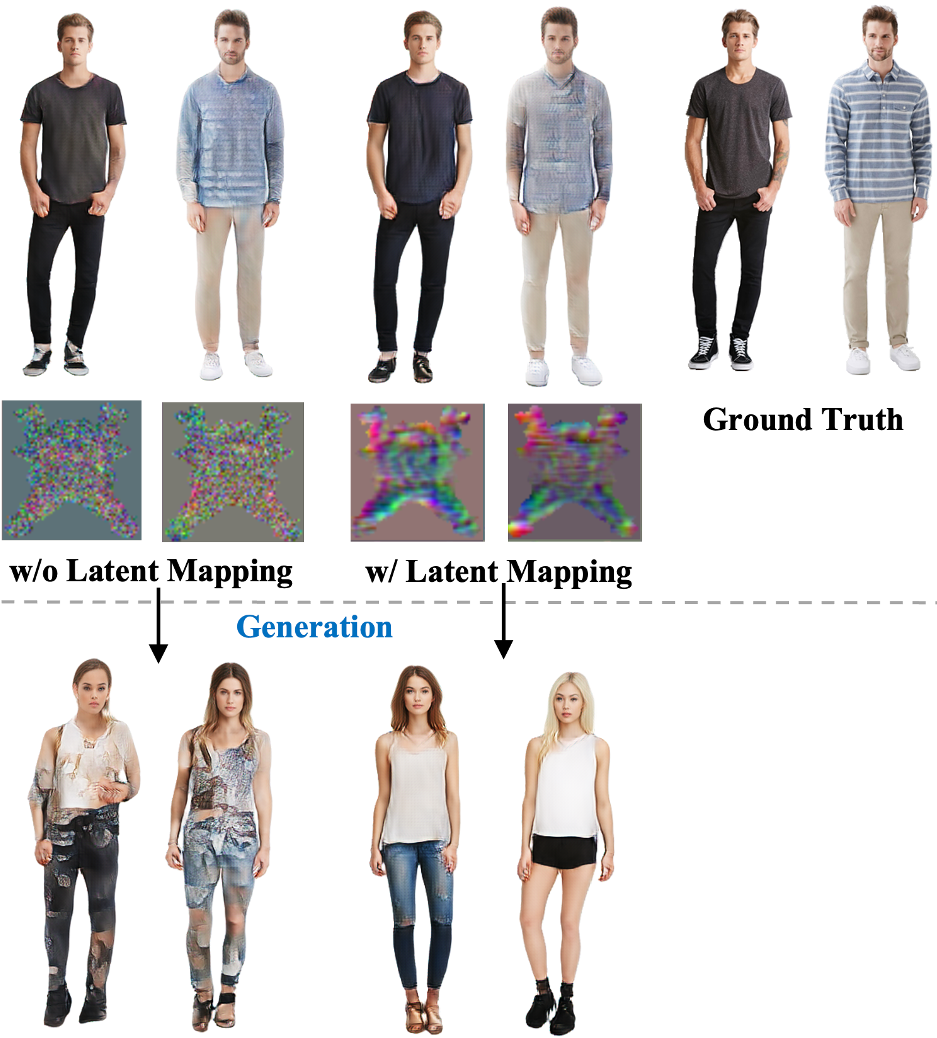}
	\end{center}
 	\vspace{-0.06in}
	\caption{Trade-off between reconstruction fidelity and generation quality on DeepFashion \cite{Liu2016DeepFashion}. Row 1: reconstruction results by auto-decoder. Row 2: latent visualization by Principal Component Analysis (PCA). Row 3: generation results.}
	\label{fig:sup_df_latent}
	\vspace{-0.12in}
\end{figure}
}

\newcommand{\figSupRenderPeople}{
\begin{figure*}[h]
	\begin{center}
        \includegraphics[width=\linewidth]{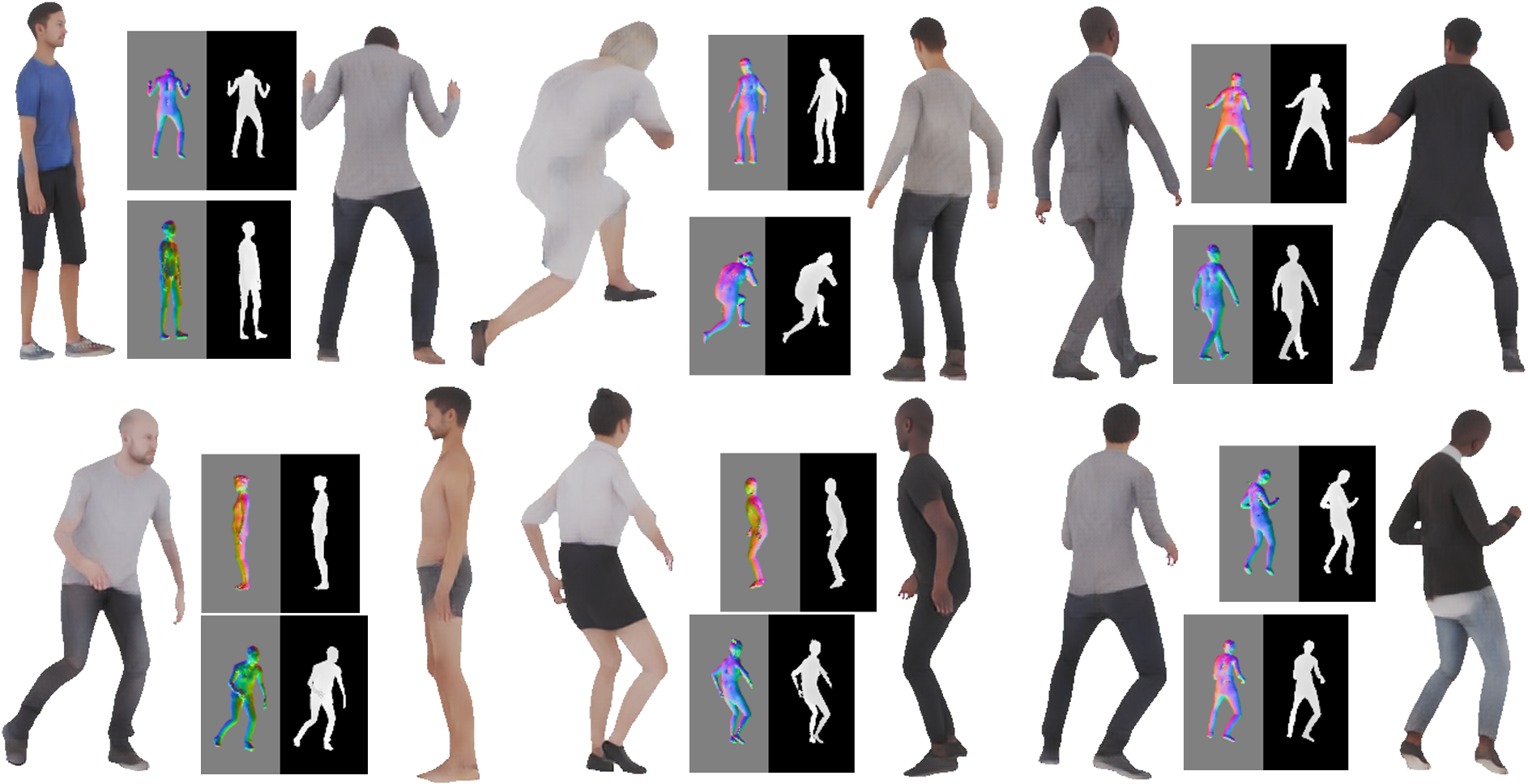}
	\end{center}
 	\vspace{-0.06in}
	\caption{Qualitative results on RenderPeople. For our method, normal and depth maps at $128\times64$ resolution are rendered by our volumetric renderer.}
	\label{fig:sup_renderpeople}
	\vspace{-0.12in}
\end{figure*}
}

\newcommand{\figVideoInv}{
\begin{figure}[t]
	\begin{center}
        \includegraphics[width=.7\linewidth]{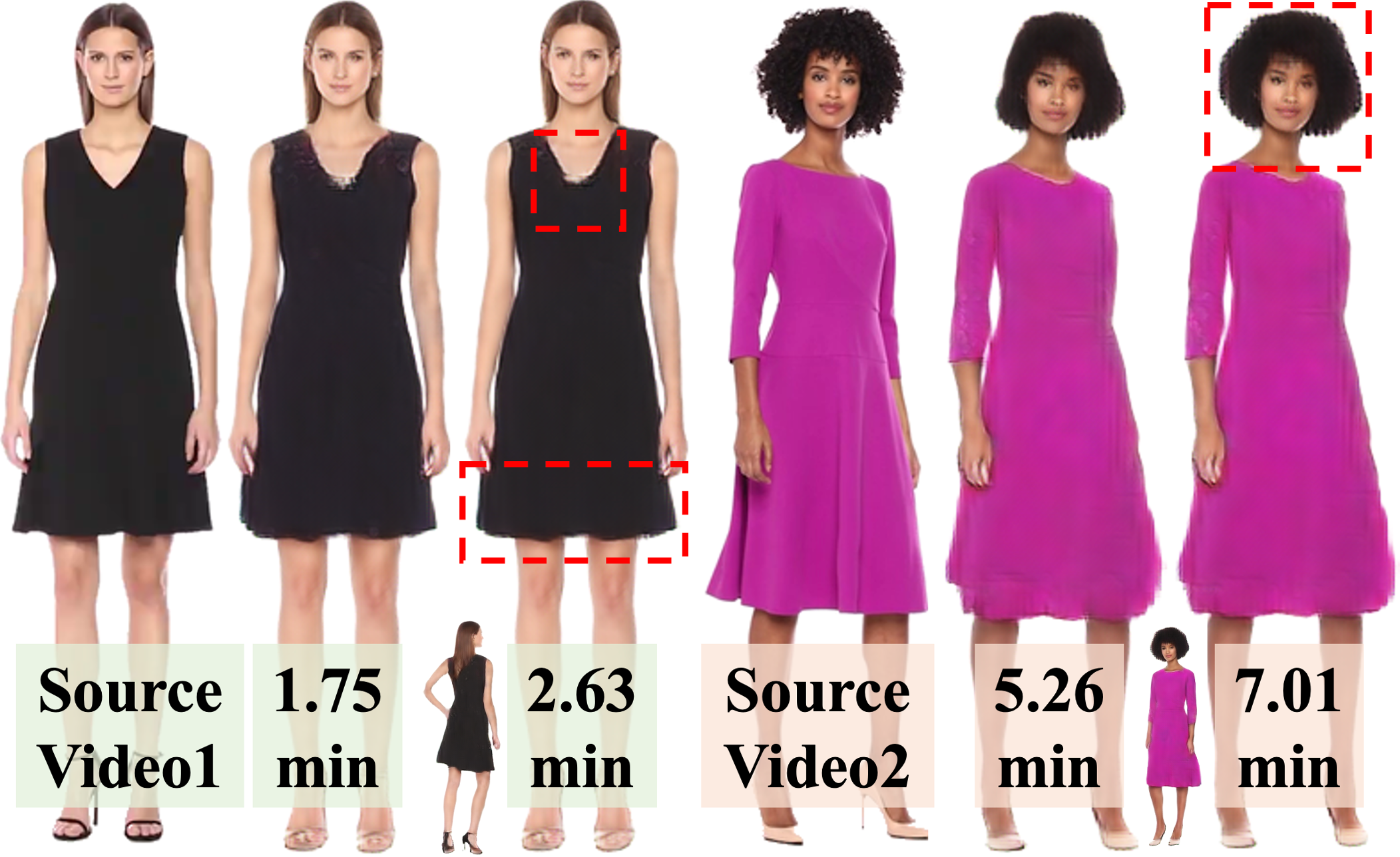}
	\end{center}
 	\vspace{-0.06in}
	\caption{Video inversion efficiency.}
	\label{fig:videoInv}
	\vspace{-0.12in}
\end{figure}
}

\section{Related Work}

\noindent{}\textbf{2D Human Generation.} Generative adversarial networks (GAN)~\cite{goodfellow2020generative} have been a great success in human faces generation~\cite{stylegan1, stylegan2, stylegan3}. However, due to the complexity of human poses and appearances, it is still challenging for GANs to generate realistic human images~\cite{humangan, tryongan, sarkar2021style, jiang2022text2human,egorend}. Scaling up the dataset has been proven effective in improving human generation quality~\cite{styleganhuman, fruhstuck2022insetgan}. Recent advancements in diffusion models~\cite{ddpm, improveddiffusion, ldm} have inspired its application in human image generation~\cite{controlnet, liu2023hyperhuman}. \cite{wang2023disco,dreampose_2023} both take human image and pose as input and utilize a pre-trained diffusion model for conditional generation, whereas we train a 3D diffusion model from scratch for unconditional generation.

\noindent{}\textbf{3D Human Generation.} Using real-scanned 3D human dataset, gDNA learns to generate detailed geometry~\cite{chen2022gdna}. With the success of 3D-aware GANs~\cite{pigan, giraffe, eg3d}, the second line of work proposes to learn 3D human generation from 2D human image collections.  
ENARF-GAN~\cite{enarfgan} is the first to combine human neural radiance fields with 3D-aware adversarial training. Super-resolution modules can be further used to increase the generation resolution~\cite{eg3dhuman, avatargen}. EVA3D~\cite{eva} proposes to use compositional human representation to increase the raw resolution of neural rendering. AG3D~\cite{ag3d} further improves the result with face and normal discriminators.   \cite{abdal2023gaussian} utilizes Gaussian Splatting \cite{kerbl3Dgaussians,Zwicker2002EWAS} for fast rendering. PrimDiff \cite{chen2023primdiffusion} parameterizes humans with volumetric primitives \cite{mvp}, and learns a diffusion model for unconditional generation. Recent advances in text-image joint distribution learning~\cite{clip, ldm} have enabled text-driven 3D generation~\cite{dreamfields, dreamfusion}. Combined with 3D human representations, open-vocabulary text-driven 3D human generation can be achieved~\cite{avatarclip, cao2023dreamavatar, cao2023guide3d}.

\figFramework

\noindent{}\textbf{Diffusion Models for 3D Generation.} Diffusion models have shown great ability in modeling complex distributions. Many 3D diffusion models have been explored in recent years based on different 3D representations, such as explicit 3D representations of point clouds~\cite{nichol2022point, luo2021diffusion, zeng2022lion} and voxels~\cite{zhou20213d, muller2022diffrf}, implicit functions \cite{jun2023shap}, triplanes~\cite{eg3d,shue20223d, wang2022rodin, gu2023nerfdiff, gupta20233dgen}, volumetric primitives \cite{mvp,chen2023primdiffusion}.

\noindent{}\textbf{High-Dimensional Structured Representation.} Existing methods \cite{eg3d,stylepeople,Chen2022UVVF,Sun2022Next3DGN,enarf,eg3dhuman,enarfgan} all rely on an implicit mapping network (\eg,  StyleGAN \cite{karras2019style}) to map 1D embedding to a triplane or UV latent in a high-dimensional space, while they still sample 1D noises in 1D space for object or human generations. Instead, we explicitly model the high-dimensional structured latent space without relying on mapping networks, which faithfully preserves the fidelity and semantic structures of the latent or embedding space. We illustrate that though the latent mapping enables a smoother latent space for diffusion learning, it degrades the reconstruction fidelity in the supp. mat. 
\section{Our Approach}

\nickname{} is a two-stage approach. In the first stage, we learn an auto-decoder containing a set of structured embeddings $\mathcal{Z}$ corresponding to the human subjects in the training dataset, and both the auto-decoder and embeddings are optimized to render pose- and view-conditioned human images with 2D supervision from the training images. The embeddings $\mathcal{Z}$ are then employed to train a latent diffusion model operating in the compact structured latent space in the second stage, which enables diverse and realistic human generations. The full pipeline is depicted in Fig. \ref{fig:framework}.

We first present the structured latent representation (Sec. \ref{method:structlat}), and then describe the auto-decoding architecture and training procedure (Sec. \ref{method:autodecoding}, \ref{method:adtrain}). Finally, we provide details for the training and sampling of diffusion in the structured latent space (Sec. \ref{method:diff}).

\subsection{Structured 3D Human Representation} \label{method:structlat}

Most existing 2D/3D human generative approaches \cite{eva,ag3d,styleganhuman,enarf,gnarf} model human appearances (\eg, clothing style) in a compact 1D latent space, ignoring the articulated structure and the semantics of the human body. Instead, we explore the articulated structure of the human body and propose to model humans on the dense surface manifold of a parametric human body. The 3D human is recorded on a 2D latent map $\mathbb{R}^{U\times V \times C}$ in the UV space of SMPL mesh \cite{smpl}. The 2D latent preserves the rich semantics and structures of human body. It is proven to capture the fine details of human appearance better than the existing 1D latent (Sec. \ref{sec:ab_latent}).  Note that different from \cite{Chan2021EfficientG3,stylepeople,ag3d} that rely on an implicit mapping network of StyleGAN \cite{karras2019style},
we explicitly model the structured latent space with explainable differentiable rendering, which faithfully preserves the fidelity and semantic structures of the sampling space. Besides, ours is distinguished from StylePeople \cite{stylepeople} by learning high-quality 3D view-consistent generations.

In addition, instead of using the discontinuous SMPL UV mapping \cite{smpl,scale,pop,Remelli2022DrivableVA,hvtr,hvtrpp,surmo} where each body patch is placed discretely, we employ a continuous boundary-free UV mapping that maintains most of the neighboring relations on the original mesh surface and preserves the clothing structures globally with a boundary-free arrangement. The UV mapping has also been proven to be more friendly for CNN-based architecture in \cite{Zeng20203DHM} than the standard discontinuous UV mapping. Besides, in contrast to existing work employing the UV-aligned feature encoding for subject-specific reconstructions \cite{scale,pop,Remelli2022DrivableVA,hvtr,hvtrpp}, ours is distinguished by encoding various subjects for 3D generation tasks.

\subsection{Structured Auto-decoder} \label{method:autodecoding}

\noindent \textbf{Structured Local NeRFs}.
The extension from 1D to 2D latent introduces more challenges for generative models due to the square growth of the global latent space. Therefore, we propose to divide the latent space into several body parts for local human modeling, and further present a structured auto-decoder $\mathbb{F}_{\Phi}$ that consists of a set of structured NeRFs $\{F_1,..., F_k, ... F_K\}$ locally conditioned on a corresponding body patch in the structured latent space as shown in Fig. \ref{fig:framework}. Specifically, each body part is parameterized by a local NeRF ${F_k}$, which model body in a local volume box $\{{b}_{min}^{k}, {b}_{max}^{k}\}$.  

To render the observation space with the estimated posed SMPL mesh $M({\beta},{\theta})$ and camera parameters of a specific identity, we query the corresponding identity latent ${z} \in \mathcal{Z}$ for all sampled points along camera rays and integrate it into a 2D feature map. To be more specific, given a 3D query point ${x}_i$ in the posed space, we first transform it to a canonical space using inverse linear blend skinning (LBS) which yields ${\hat{x}}_i$. Then in the canonical space, we find the nearest face $f_i$ of the SMPL mesh for the query point ${\hat{x}}_i$ and $(u_i, v_i)$ are the barycentric coordinates of the nearest point on $f_i$.  We then obtain the local UV-aligned feature ${z}_i = B_{u_i,v_i}({z})$ of the query point $x_i$, where $B$ is the barycentric interpolation function and ${z}$ is the 2D structured latent. Given camera direction ${d}_i$, the appearance features ${c}_i$ and density ${\sigma}_i$ of point ${x}_i$ are predicted by
\begin{equation} \label{eq:windowfunction}
    \vspace{-0.06in}
    \{{c}_i, {\sigma}_i\} = \frac{1}{\sum\omega_a}\sum_{a\in\mathbb{A}_i}\omega_a F_k({{\hat{x}}_i^k, {d}_i, {z}_i)},
    \vspace{-0.02in}
\end{equation}
\noindent where $\mathbb{A}_i$ indicates the NeRF sets where point ${x}_i$ falls in, ${\hat{x}}_i^k$ is the local coordinate of ${x}_i$ in NeRF $F_k$, and $\omega_a = \text{exp}(-m(\hat{{x}}_i^k({x})^n + \hat{{x}}_i^k(y)^n + \hat{{x}}_i^k(z)^n))$ is blending weight when ${x}_i$ falls in multiple volume boxes. $m$ and $n$ are chosen empirically.

Note that each local NeRF $F_k$ is conditioned on a local UV-aligned feature $z_i$ by $F_k({{\hat{x}}_i^k, {d}_i, {z}_i)}$. We integrate all the radiance features of sampled points into a 2D feature map ${I_F} \in \mathbb{R}^{H\times W \times \bar{C}}$ through volume renderer $G_1$ \cite{Kajiya1984RayTV}:
\begin{equation} \label{eq:reslow}
    \vspace{-0.06in}
    {I_F} = G_1({z}, {\beta}, {\theta}, cam; \mathbb{F}_{\Phi}).
    \vspace{-0.02in}
\end{equation}

The structured representation inherits human priors and renders human images in a canonical space to disentangle pose and appearance learning, and enables adaptive allocation of network parameters for efficient training and rendering. In contrast to \cite{zheng2022structured} that parameterizes the human body with hundreds of vertex-level NeRFs, we adopt the part-level structure of EVA3D \cite{eva} with 16 local NeRFs for smooth clothing style mixing. In contrast to EVA3D, we use a lightweight auto-decoder architecture, conditioned on the structured 2D latent.

\noindent \textbf{Efficient Geometry-Aware Global Style Mixer}. The compositional NeRFs model each body part separately in a canonical space, while they struggle to learn the full-body appearance style as a whole, especially for dress. In addition, the compositional volume boxes with fixed sizes are not effective in reconstructing loose clothing, \ie, predicting hemline or between-leg offsets of dress, as observed in \cite{eva,ag3d}. To solve these issues, we propose a global style mixer $G_2$ that learns to mix the compositional features of each body part to learn full-body appearance style in the compact image space.
\begin{equation}
    \vspace{-0.06in}
    {I^{+}_{RGB}}=G_2 \circ G_1({z}, {\beta}, {\theta}, cam; \mathbb{F}_{\Phi}).
    \vspace{-0.02in}
\end{equation} 

The style mixer is built upon Transposed CNN \cite{Shelhamer2014FullyCN}. $G_2$ mixes neighbor 4 pixels in the feature map ${I_F}$ by a receptive field of 4, and upsamples the feature map with $4\times$ super-resolution in image space, which enables efficient geometry-aware rendering as used in  ~\cite{Chan2021EfficientG3}. Refer to more details in the supp. mat. 

\subsection{Joint Learning of Auto-decoder} \label{method:adtrain}
\nickname{} is trained to optimize renderers $G_1$, $G_2$ and structured embeddings $\mathcal{Z}$. We employ 1) Reconstruction Loss including Pixel Loss, Perceptual Loss, Face Identity Loss, and Volume Rendering Loss, 2) Adversarial Loss, and 3) Regularization Loss for the learning of geometry and embeddings $\mathcal{Z}$ in training, with Adam \cite{adam} as the optimizer. Refer to the supp. mat. for more details.


\subsection{Structured Latent Diffusion Model} \label{method:diff}

After embedding training subjects in a structure 2D latent space, we trained a latent diffusion model~\cite{ldm} to learn to sample in this space. Diffusion models~\cite{ddpm} are probabilistic models that learn a data distribution $z \sim p_z$ by gradually removing noises from random Gaussian noises. Specifically, the noise-removing process corresponds to a reverse Markov Chain of length $T$. For each step $t$, the model, parameterized by U-Net $\epsilon_\theta$, learns to predict the noise $\epsilon$ from the input noised sample $z_t$. The training objective can be formulated as
\begin{equation}
    L_{D} = \mathbb{E}_{\epsilon \sim \mathcal{N}(0,1), t \sim \mathcal{U}(1, T)}\left[ \| \epsilon - \epsilon_\theta(z_t, t) \|_2^2 \right].
\end{equation}
The structured 2D latents are spatially aligned with the UV space, and we further improve the diffusion model learning by using structure-aligned normalization. Specifically, we calculate the mean and variance of each UV pixel across the whole dataset and normalize latent $z$ for each pixel independently.

\figCmpUBC

\figQtThuDf

\figUsCmpPd

\tabQt

\section{Experiments}

\subsection{Experimental Setup}
\noindent \textbf{Datasets}. We perform experiments on three datasets: UBCFashion~\cite{ubcfashion} (500 monocular human videos with natural fashion motions), THuman 2.0~\cite{thuman2} (526 human scans), and RenderPeople~\cite{renderpeople} (796 high-quality 3D human models with diverse identities and clothes). We render each 3D scan/model of THuman 2.0 and RenderPeople into multi-view images for training.

\noindent \textbf{Metrics}. We measure the diversity and quality of generated human images using the Frechet Inception Distance (FID)~\citep{heusel2017gans} between 50k generated images and real images at the resolution of $512\times 512$. In addition, we conduct a perceptual user study and report how often the generated images by our method are preferred over other methods in terms of both overall appearance quality and face quality.

\noindent \textbf{Baseline Methods}. EG3D~\cite{eg3d} and StyleSDF \cite{stylesdf}, EVA3D~\cite{eva}, AG3D~\cite{ag3d} and PrimDiff \cite{chen2023primdiffusion}  are the state-of-the-art methods for 3D-aware generation of static objects and articulated 3D humans respectively.

\subsection{Comparisons to SOTA Methods}

\noindent \textbf{Quantitative Comparisons}. Table. \ref{tab:cmpquant} summarizes the quantitative comparisons against SOTA methods. Our approach achieves better results than all four 3D-aware GANs on all datasets in terms of FID. Our method outperforms others by a significant margin on RenderPeople and THUman2.0, and the performance of EVA3D on THUman2.0 is much worse partly because of the small pose variance of THUman2.0. The improvement is also confirmed by our user study in Fig. \ref{fig:userstudy}. More than 20 participants are asked to select the images with better overall quality or face quality from the random generations of AG3D and ours. It was observed that about 74.68\% and 72.5\% of generated images by our method are considered to be more realistic in terms of overall quality and face quality respectively. More metrics discussion can be found in the supp. mat.

\noindent \textbf{Qualitative Comparisons}. The improvements over baselines are further confirmed qualitatively in Fig. \ref{fig:cmp_ubc}, where we show the renderings of each approach on UBCFashion. Compared to 3D-aware GANs, \nickname{} is capable of generating diverse humans with various clothing styles and different skin colors with powerful diffusion-based sampling. While GAN-based sampling lacks such diversity. In addition, Fig. \ref{fig:cmp_ubc} shows that our method generates view-consistent humans with high-quality appearances and details (\eg, high heels) under different poses and views in different clothing styles including dresses and even for challenging hairstyles, where the consistency is well-preserved in the structured latent space.

The qualitative comparisons on THUman2.0 against EVA3D are shown in Fig. \ref{fig:sup_thuman}. We can even learn to generate realistic human images with reasonable geometry reconstructions (\eg, normal, depth) from single images on DeepFashion~\cite{Liu2016DeepFashion} as shown in Fig. \ref{fig:df}. Refer to more details in the supp. mat.

The qualitative comparisons on RenderPeople \cite{renderpeople} are shown in Fig. \ref{fig:cmp_rp}, where the learned geometries are  visualized as normal and depth maps. Besides consecutive monocular sequences, our method is also capable of learning 3D diffusion models from multi-view images with several static poses, and ours significantly outperforms existing SOTA 3D-aware GANs in diverse generations. We synthesize high-quality faces \cnum{3}\cnum{4}\cnum{5} with adversarial training while PrimDiff \cite{chen2023primdiffusion} cannot produce the same level of realism. Ours also generalizes well to different poses, including some extreme poses.

    \textbf{Texture Transfer}. Both PrimDiff (PD) and ours support texture transfer by editing UV latents, \ie, combining \cnum{1} and \cnum{2} shown in Fig. \ref{fig:cmpPd}. \textbf{PrimDiff renders the combined latent directly in a} \textbf{decoder-free way}, which leads to artifacts. However, ours has an auto-decoder with a {global style mixer} ($G_2$) to decode the combined latent, which enables better style mixing \cnum{3}\cnum{4}.

\noindent \textbf{Efficiency}. Similar to most latent diffusion models {{\eg, Stable Diffusion \cite{ldm}}}, ours is not trained end-to-end. It takes about 3.5 days to train an auto-decoder from about 80K images on UBCFashion and 3 days for latent diffusion on 4 NVIDIA V100 GPUs, more efficient than EVA3D \cite{eva} (5 days on {8} V100). In inference, our rendering network runs at 9.17 FPS to render $512^2$ resolution images on a V100 GPU, $1.94\times$ faster than AG3D. It takes about {127.55 s} to sample 64 latents using DDIM \cite{song2020denoising} with 100 steps on one V100 GPU.

\subsection{Ablation Study}
\noindent \textbf{Auto-decoder: 2D vs. 1D Latent for Human Reconstruction}. \label{sec:ab_latent}
We compare the performance of our structured 2D latent with the widely adopted global 1D latent for human reconstructions in auto-decoder both quantitatively and qualitatively in Fig. \ref{fig:ab_ad_quan_latent} and Fig. \ref{fig:ab_ad_qual}. The reconstruction quality measured by LPIPS \cite{lpip} on 4K samples of UBCFashion is reported in Fig. \ref{fig:ab_ad_quan_latent}. No improvement is observed for 1D latent when increasing the latent size from 128 to 4096. Whereas, our structured 2D latent outperforms the 1D latent by a large margin in reconstruction. And the quality can be significantly improved by increasing the resolutions or channels of the 2D latent. Note that our structured 2D latent also works at extremely low resolutions, \eg, $16\times16\times8$, which outperforms the 1D latent of 2048 with the same amount of parameters. Note that though $64\times64\times32$ achieves comparable performances with $128\times128\times16$, the latter is more friendly for editing tasks with higher resolutions.

\tabAblation


In addition, the qualitative results in Fig. \ref{fig:ab_ad_qual} suggest that 1D latent fails to capture detailed high-quality details for face or cloth patterns, and often generates blurry rendering results. This is because the global 1D latent neither encodes semantics nor structure features for the articulated human body. In contrast, our 2D latent  captures the face structures and clothing patterns under the same reconstruction supervision (more details in the supp. mat). The comparisons in Fig. \ref{fig:ab_ad_qual} are based on the best performances of 1D latent (256) and 2D latent ($128\times128\times16$) according to the test results in Fig. \ref{fig:ab_ad_quan_latent}.

\noindent \textbf{Auto-decoder: Adversarial Training}. Different from the training on rigid objects with only reconstruction losses as supervision \cite{ntavelis2023autodecoding}, rendering articulated humans with complicated clothing styles is far more challenging. In addition, per-pixel reconstruction losses are often sensitive to misalignment caused by human pose or camera estimation errors, especially for single-view setup. Instead, a discriminator with adversarial training is employed in our auto-decoder framework to enforce realistic rendering with high robustness to pose or camera estimation errors. The adversarial training enables both high-fidelity and high-quality image reconstruction, as confirmed in Fig. \ref{fig:ab_ad_qual}.

\noindent \textbf{Diffusion: Structure-aligned Normalization}. We also analyze the normalization methods for our structured latent learning in diffusion quantitatively and qualitatively in Tab. \ref{tab:ab_diff_quant} and Fig. \ref{fig:ab_diff_norm}. Thanks to the structured alignment of the latent space and the auto-decoder, the learned latent space is well-structured and enables the training of diffusion even without normalization ('None'). However, Tab. \ref{tab:ab_diff_quant} illustrates that this is non-trivial since the well-adopted standard normalization even prohibits 3D structured diffusion. Instead, a unique structure-aligned normalization (see Sec. \ref{method:diff}) further improves the generation results quantitatively. Moreover, Fig. \ref{fig:ab_diff_norm} suggests that the proposed structure-aligned normalization reduces the distance between the latent distribution learned in auto-decoder and normal distribution. Furthermore, an illustration of the structure-aligned normalization is shown as \cnum{1} where the standard deviation and mean for each pixel are shown after normalization, which illustrates that our 2D latent encodes the structure information (\eg, symmetry) of the human body in differentiable rendering without explicit latent structure supervision.

\noindent \textbf{Latent Diffusion}. We also analyze the effect of latent diffusion in a challenging compositional generation task in Fig. \ref{fig:sup_stylemixing}. Refer to Sec. \ref{sec:apps} for more details. 


\figCtrlPart

\subsection{Controllable Human Generation and Editing}
\label{sec:apps}
Emerging from the technical choice of 2D latent and diffusion model, we show various human generation and editing applications below, which would potentially boost the productivity of fashion designers.

\noindent {\textbf{Pose-view-shape Control}}. Benefiting from the articulated human representation, Struct-LDM enables designers to freely control the generations under different {pose, view, and shape} conditions as shown in Fig. \ref{fig:ctrl}.

\noindent {\textbf{Interpolation}}. As shown in Fig. \ref{fig:ctrl} d), we interpolate two latent codes in diffusion~\cite{song2020denoising} to generate a smooth transition. Though only 500 identities of UBCFashion are used for training, semantically meaningful 2D latent space can be learned by our auto-decoder and diffusion model.

\noindent \textbf{Compositional Generations}. As shown in Fig.~\ref{fig:partediting} b),
taking six body parts from six generated source identities in a), \nickname{} is capable of blending these parts in the unified structure-aligned latent space and using a {Diff-Render} procedure for decent style mixing. It includes latent noising and denoising steps by part-aware diffusion, and a rendering step for decent style mixing. Geometry styles (\eg, neckline (1), cuff (2), hemline (3)) are well transferred with high fidelity, and different skin colors are also decently blended, \ie, \cnum{2} to (1). Note that the skin and skirt colors are different for the compositional generations of \cnum{1}\cnum{2}\cnum{3}\cnum{4}\textcolor{green}{\cnum{5}} and \cnum{1}\cnum{2}\cnum{3}\cnum{4}\textcolor{blue}{\cnum{6}} because of our full-body style mixing strategy.

\textbf{The Effect of Diff-Render.} In Fig. \ref{fig:sup_stylemixing}, we analyze the effect of latent diffusion in {Diff-Render} in a more challenging compositional generation task, which suggests that the auto-decoder alone fails to blend the pink skirt and the black skirt, \eg, `w/o Diff'. In contrast, a procedure of latent noising and denoising in diffusion enables plausible color blending with a proper setup of steps $S$ or noise factor $\eta$. The diffusion is based on DDIM \cite{song2020denoising} sampling.

\figSupStyleMix

We further detail the effect of {Diff-Render} in Fig. \ref{fig:part_diffusion}. When we mix different styles, including clothing geometry style and skin or clothing colors, we would like to faithfully preserve the geometry style of each source part and instead blend different colors in a decent way. Yet directly blending the colors, \eg, \cnum{1}, often leads to artifacts shown by 'w/o Diff' in Fig. \ref{fig:part_diffusion}. Though the effect can be alleviated by latent noising and denoising that yields a better color blending with more denoising steps, the fidelity of clothing shape also degrades, \eg, \cnum{2}\cnum{3}\cnum{4}. Instead, we employ part-aware diffusion to solve the local inconsistencies in skin tone by utilizing the learned diffusion prior. The part-aware diffusion mechanism (supp. mat.) allows users to freely select one specific part to locally edit or enhance the generations without losing geometry fidelity, and it also requires fewer steps to generate a desired enhancement.

\noindent {\textbf{Part-aware Editing}}. \nickname{} allows users to locally edit the generations, as shown in  Fig. \ref{fig:partediting} c). It enables applications such as identity swapping, and local clothing editing, which is achieved by the above-mentioned Diff-Render.
    
\noindent {\textbf{3D Virtual Try-on}}. As a byproduct of part-aware editing, \nickname{} supports 3D virtual try-on, \ie, rendering view-consistent humans wearing different clothes while preserving the identity, as shown in Fig. \ref{fig:partediting} c) (6) and Fig. \ref{fig:ctrl} a).

\noindent {\textbf{Full-body Style Transfer}}. In addition to part-aware editing, users are also allowed to transfer the color match of a human asset to a new identity (Fig. \ref{fig:partediting} c) (7)), which is achieved by applying Diff-Render in the full-body latent space.

\noindent {\textbf{Inversion}}. Fig. \ref{fig:appInv} shows the inversion of in-the-wild images: \cnum{1} target image, \cnum{2} inversion, \cnum{3} a new pose rendering, by a pretrained model on DeepFashion.

\figPartInv

\section{Discussion}

We propose a new paradigm for 3D human generation from 2D image collections. The key is the structured 2D latent, which allows better human modeling and editing. A structural auto-decoder and a latent diffusion model are utilized to embed and sample the structured latent space. Experiments on three human datasets show the state-of-the-art performance, and qualitative generation and editing results further demonstrate the advantages of the structured latent.

\noindent\textbf{Limitations}. \textbf{1)} We train models from scratch as in EVA3D/AG3D/ PrimDiff. The lack of a diverse in-the-wild human dataset with accurate registration is a common problem in this field. Due to the limited scale and dataset bias, diversity is not comparable to 2D diffusion models~\cite{ldm}. However, we \textbf{outperform the baselines EVA3D and AG3D in} \textbf{diversity} as shown in Fig. \ref{fig:cmp_ubc}. \textbf{2)} Limited by the auto-decoder training, it is challenging to learn from single-view 2D image collections \cite{ntavelis2023autodecoding}, \eg, DeepFashion~\cite{Liu2016DeepFashion}. However, the structured latent representation makes it possible to auto-decode 3D humans from single images on DeepFashion, as shown in Fig. \ref{fig:df}, where our method generates realistic human images with reasonable geometry reconstructions (\eg, normal, depth).

\section*{Acknowledgement}

This study is supported by the Ministry of Education, Singapore, under its MOE AcRF Tier 2 (MOET2EP20221- 0012), NTU NAP, and under the RIE2020 Industry Alignment Fund – Industry Collaboration Projects (IAF-ICP) Funding Initiative, as well as cash and in-kind contribution from the industry partner(s).




\section*{Appendix}
\appendix
\renewcommand{\thesection}{\Alph{section}}

\section{Implementation} 

\subsection{Network Architecture}
\noindent \textbf{Structured Auto-decoder.} We adopt the decoder architecture of StyleSDF and EVA3D for each structured local NeRFs. For each subnetwork, multiple MLP, and FiLM SIREN activation \cite{Sitzmann2020ImplicitNR} layers are stacked alternatively, and at the end of each subnetwork, two branches are used to separately estimate SDF value and RGB value. Different numbers of network layers for different body parts are assigned empirically: 4 layers for Head; 3 layers for Shoulder + Upper Spine, Middle Spine, Lower Spine; 2 layers for Right Upper Arm, Right Arm, Right Hand, Left Upper Arm, Left Arm, Left Hand, Right Upper Leg, Right Leg, Right Foot, Left Upper Leg, Left Leg, Left Foot with a similar design as EVA3D, whereas we adopt a more light-weight architecture with fewer layers for each subnetwork. The detailed diagram can be found in \cite{eva}. We render human features at $128\times128$ resolution by volume rendering using the structured auto-decoder.

\noindent \textbf{Global Style Mixer.} We utilize a receptive field of 4 in the experiments, \ie   upsampling the $128\times128$ renderings to $512\times512$ images. We employ two convolution blocks each containing a bilinearly upsampling step and two convolutional layers with a kernel size of 3 to upsample the images by a factor of 4.

\noindent \textbf{Latent Size.} Depending on the scale of the training dataset, the latent size is $128\times128\times16$ for UBCFashion \cite{ubcfashion}, $128\times128\times24$ for RenderPeople \cite{renderpeople}, $64\times64\times24$ for THUman2.0 \cite{thuman2}, and $64\times64\times32$ for DeepFashion \cite{Liu2016DeepFashion}.
 
\noindent \textbf{Discriminator.} We adopt the discriminator architecture of PatchGAN \cite{pix2pix,esser2020taming} for adversarial training. Note that different from EG3D \cite{eg3d} that applies the image discriminator at both resolutions, we only supervise the final rendered images with adversarial training and supervise the volumetric features with reconstruction loss.

\noindent \textbf{Diffusion Model.} Our diffusion model is based on the UNet architecture of \cite{improveddiffusion}, with four ResNet \cite{He2015DeepRL} blocks and a base channel of 128. 

\subsection{Optimization of Auto-decoder} \label{method:adtrain}
\nickname{} is trained to optimize renderers $G_1$, $G_2$ and structured embeddings $\mathcal{Z}$. Given a ground truth image $I_{gt}$, we predict a target RGB image ${{I^+_{RGB}}}$ with the following loss functions: 

\noindent \textbf{Pixel Loss}. We enforce an $\ell_1$ loss between the generated image and ground truth as ${L}_{pix} = \|I_{gt} - {I^+_{RGB}} \|_1$.

\noindent \textbf{Perceptual Loss}. Pixel loss is sensitive to image misalignment due to pose estimation errors, and we further use a perceptual loss \cite{PerceptualLosses} to measure the differences between the activations on different layers of the pre-trained VGG network \cite{vgg} of the generated image ${I^+_{RGB}}$ and ground truth image $I_{gt}$,
\begin{equation}
	{L}_{vgg}=\sum \frac{1}{N^{j}}\left\|g^{j}\left(I_{gt}\right)-g^{j}\left({I^+_{RGB}}\right)\right\|_2,
\end{equation}
\noindent where $g^j$ is the activation and $N^j$ the number of elements of the $j$-th layer in the pre-trained VGG network.

\noindent \textbf{Adversarial Loss}. We leverage a multi-scale discriminator $D$ \cite{pix2pixhd} as an adversarial loss ${L}_{adv}$ to enforce the realism of rendering, especially for the cases where estimated human poses are not well aligned with the ground truth images.

\noindent \textbf{Face Identity Loss}. We use a pre-trained network to ensure that the renderers preserve the face identity on the cropped face of the generated and ground truth image,
\begin{equation}
	{L}_{{face}}= \| N_{ {face}}\left(I_{gt}\right)-N_{ {face}}\left({I^+_{RGB}}\right)\|_2,
\end{equation}
\noindent where $N_{face}$ is the pre-trained SphereFaceNet \cite{Liu2017SphereFaceDH}.

\noindent \textbf{Volume Rendering Loss}. We supervise the training of volume rendering at low resolution, which is applied on the first three channels of ${I_F}$, ${L}_{vol} = \|{I_F}[:3] - I^D_{gt}\|_2$. $I^D_{gt}$ is the downsampled reference image. 

\noindent \textbf{Latent Regularization}. To allow better learning of the latent diffusion model, we regularize the latent with L2 regularization and TV loss.

\noindent \textbf{Geometry Regularization Loss}. Following EVA3D~\cite{eva}, we predict the delta signed distance function (SDF) to the body template mesh. Therefore, we penalize the derivation of delta SDF predictions to zero $\mathcal{L}_{\text{eik}} = {E}_{{x}}[\|\nabla (\Delta d({x})) \|_2^2]$~\cite{gropp2020implicit}.

The networks were trained using the Adam optimizer \cite{adam}. It takes about 3.5 days to train an auto-decoder from about 80K images on UBCFashion on 4 NVIDIA V100 GPUs, and about 3 days to train a diffusion model (based on \cite{improveddiffusion}) with 4 NVIDIA V100 GPUs.

\subsection{Training Data Precessing} UBCFashion \cite{ubcfashion} contains 500 sequences of fashion videos, and we uniformly extract about 80K images from these videos as training data. We render 24 multi-view images for RenderPeople \cite{renderpeople} and THUman2.0 \cite{thuman2}, which yield about 190K images and 12K images separately. For DeepFashion \cite{Liu2016DeepFashion}, we directly use the pre-processed subset with 8K images from EVA3D \cite{eva} as our training data. We crop and resize the images to $512\times512$ for training.

\figDfLatent

\subsection{Part-aware Diffusion}
Benefiting from the semantic design of latent space, \nickname{} supports local editing by {part-aware diffusion} in inference. Given a part mask $\bm{M}\in [0,1]$ and reference latent $\bm{y}_{0}$, we generate $\bm{x}_{t-1}$ from noised $\bm{x}_{t}$ based on the estimated value for $\bm{x}_{0}$, namely $\bar{\bm{x}}_0^{(t)}$, which is refined to get $\hat{\bm{x}}_0^{(t)}$: 
\begin{equation}
   {\min_{\hat{\bm{x}}_0^{(t)}} \left\Vert \hat{\bm{x}}_0^{(t)} - \bar{\bm{x}}_0^{(t)} \right\Vert_2 + \lambda \left\Vert (\bm{1} - \bm{M}) \odot \left(\bm{y}_0 - \bar{\bm{x}}_0^{(t)}\right) \right\Vert_2}
   \label{eq:partdiff1}
\end{equation}
where $\odot$ denotes the Hadamard product, $\lambda=0.5$ is a hyper-parameter controlling the balance between the diffusion prior and the degradation constraint. It admits a closed-form solution: 
\begin{equation}
 {\hat{\bm{x}}_0^{(t)} = \frac{\bar{\bm{x}}_0^{(t)} + \lambda(\bm{1}-\bm{M})^2\odot\bm{y}_0}{\bm{1} + \lambda(\bm{1}-\bm{M})^2}}
 \label{eq:partdiff2}
 \end{equation}
which can be optimized using SGD when a closed-form solution is not feasible. The part-aware latent diffusion is similar to image inpainting in  \cite{Yue2022DifFaceBF}. All the mathematical operations are pixel-wise.
 
\section{Additional Experimental Results}


\subsection{The Effect of Latent Mapping for  Reconstruction and Generation}

Autodecoding human latents from single images is challenging due to sparse observations. Fig. \ref{fig:sup_df_latent} shows the reconstruction results, latent visualizations, and generation results. It is observed that the learned latents are noisy and unfriendly for latent diffusion, i.e., the generations are noisy. Instead, we employ a mapping network to smooth the unobserved body parts in the latent space, which yields smoother latents and enables realistic generations. Though the latent mapping improves the generation quality, it degrades the reconstruction fidelity, which imposes challenges of learning auto-decoders from single images. The mapping network consists of 3 convolutional layers with a kernel size of 5, whereas the mapping network is not required for datasets of video sequences or multi-view images such as UBCFashion, RenderPeople, and THUman2.0.

\tabQtDF

\subsection{Experimental Results on DeepFashion}

\noindent \textbf{Qualitative and Quantitative Results.} Existing auto-decoder-based methods generally require multi-view images or video sequences of objects to train an auto-decoder  \cite{ntavelis2023autodecoding}. Instead, with the structured latent representation, we show that it is even possible to auto-decode 3D humans from single images on DeepFashion \cite{Liu2016DeepFashion}, as shown in Fig.6 of the paper, where our method generates realistic human images with reasonable geometry reconstructions (e.g., normal, depth). The quantitative results are shown in Tab. \ref{tab:quant_df}. Our method outperforms existing 3D GAN methods, including StyleSDF \cite{stylesdf}, EG3D \cite{eg3d}, ENARF-GAN \cite{enarfgan}, and achieve comparable results as the publicly released EVA3D. However, the best performances of AG3D and EVA3D achieve lower FID. 

\label{sec:df}

\figEditWild
\subsection{Editing in-the-wild images.} As shown in Fig. \ref{fig:sup_edit_wild}, to edit in-the-wild (cross-dataset) Internet images \cnum{1}, an inversion step \cnum{2} is performed, and images are edited via part-aware diffusion: \cnum{3} a new pose rendering, \cnum{4} new identity, \cnum{5} shoes, \cnum{6} T-shirts, \cnum{7} pants. Part-aware editing can also be applied for generated humans \cnum{8}\cnum{9}, i.e., transferring the style of the Internet images (T-shirt \cnum{8}), and editing the shoes \cnum{9}.

\figVideoInv
\subsection{Inversion Efficiency} Similar to most latent diffusion models \cite{ldm}, ours is not trained end-to-end. However, our training (\textbf{3.5 days} auto-decoder + \textbf{3 days} diffusion on \textbf{4} V100 GPUs) is more efficient than 3D-aware GAN EVA3D (5 days on \textbf{8} V100). Once trained, a new video sample can be added by inversion as shown in Fig. \ref{fig:videoInv}. Source video 1 and 2, with 110 frames each, take \textbf{1.75 min} and \textbf{7.01 min} (for complexity in hair) on 8 V100 to inverse respectively.

\figSupRenderPeople

\section{Future Work}
As discussed in Sec. \ref{sec:df}, our framework is capable of autodecoding 3D human latents from single images, and future work would be to improve the performances on DeepFashion.

\renewcommand{\bibname}{\protect\leftline{References}}
\bibliographystyle{splncs04}
\bibliography{main_short}

\end{document}